\documentclass{article} 
\usepackage{iclr2027_conference,times}

\usepackage{amsmath,amsfonts,bm}

\def\eqref#1{equation~\ref{#1}}

\def\1{\bm{1}}

\DeclareMathAlphabet{\mathsfit}{\encodingdefault}{\sfdefault}{m}{sl}
\SetMathAlphabet{\mathsfit}{bold}{\encodingdefault}{\sfdefault}{bx}{n}

\usepackage{hyperref}
\usepackage{url}
\usepackage{amssymb}

\usepackage{algorithm}
\usepackage{algpseudocode}
\usepackage{graphicx}
\usepackage{multirow}

\hypersetup{
hidelinks,
    pdftitle={From Grammar Deltas to Token Masks:
    Incremental Constrained Decoding under Runtime Grammar Evolution},
    pdfauthor={Hantao Hua, Jiming Su, Yonggang Che, Feng Zhu}
}

\title{From Grammar Deltas to Token Masks: \\ Incremental Constrained Decoding \\ under Runtime Grammar Evolution}

\author{Hantao Hua\textsuperscript{1}, Jiming Su\textsuperscript{2}, Yonggang Che\textsuperscript{1} \& Feng Zhu\textsuperscript{2} \\
\textsuperscript{1}College of Computer Science and Technology, National University of Defense Technology\\
\textsuperscript{2}College of Systems Engineering, National University of Defense Technology\\
\texttt{\{ht\_hua,zhufeng\}@nudt.edu.cn} \\
}

\iclrfinalcopy 
\begin{document}

\maketitle

\begin{abstract}
Grammar-constrained decoding enables large language models to generate outputs that satisfy formal constraints, but in code and domain-specific language generation, the relevant constraints are not always fully available before decoding and may emerge progressively from new declarations, inferred type information, or runtime analysis feedback. Existing structured-decoding engines efficiently handle fixed, predeclared, or previously seen constraint structures; however, when a first-seen grammar update changes the parser state induced by an already committed prefix, the straightforward solution reconstructs the constraint state under the updated grammar, repeating parsing work already performed over the prefix. We introduce \textbf{GramDelta}, an incremental framework for constrained decoding with append-only runtime grammar updates. GramDelta maintains parse state over the committed prefix and propagates only changes induced by newly added grammar rules through historical dependencies, avoiding prefix replay. It then maps the updated parse state to exact next-token masks via compilation-free factorized projection over shared historical charts, avoiding per-update mask-cache recompilation. GramDelta therefore preserves the result of full reconstruction while making update cost sensitive to the actual impact of a grammar change. Experiments across four model configurations show that GramDelta adds negligible overhead without updates; under sustained runtime grammar evolution, it reduces median first-seen grammar-update latency by $10.7$--$11.5\times$ and reduces the additional TPOT overhead over unconstrained decoding by $\mathbf{53.1}\text{--}\mathbf{81.9}\%$ relative to the strongest reconstruction baseline. These results enable efficient constrained decoding when constraints progressively emerge and evolve with the generated prefix.
\end{abstract}

\section{Introduction}
Grammar-constrained decoding \citep{geng2023grammar,scholak-etal-2021-picard,dong2025xgrammar,hua2026gramtoken} enables large language models (LLMs) to reliably generate structured outputs that satisfy formal constraints by restricting invalid tokens during decoding.
However, in code and domain-specific language generation, the effective
constraints are not always fully determined before decoding begins and may
instead emerge progressively as generation proceeds \citep{li2026xgrammar}.
A newly generated declaration can introduce identifiers and scopes that did
not previously exist \citep{zhang2026decode}; type inference can further restrict admissible
expressions \citep{mundler2025type}; and static or runtime analyses \citep{li2025correctness} can determine which fields, APIs, or program actions are valid for the current partial program.
The common feature of these settings is that the information determining valid continuations becomes available only as the generation context is established.

Recent work supports such runtime-dependent constraints from several
directions.
Structured-generation engines can dispatch among dynamically selected grammar
components and reuse computation across related constraint structures
\citep{li2026xgrammar}, while code-generation methods can instantiate or
derive constraints from an evolving program environment
\citep{mundler2025type,zhang2026decode}.
These systems show that constrained decoding need not operate over a language
whose effective structure is fully fixed before generation.

However, existing methods primarily address how to
select, construct, or apply runtime constraints, rather
than how to update the parser state already computed
for a committed prefix. When a first-seen grammar
delta is introduced, newly added productions may
change the parsing results of this unchanged prefix.
Full reconstruction therefore reprocesses historical
input, including computations unaffected by the update.

To avoid this redundant work, we propose \textbf{GramDelta}, an incremental
constrained-decoding framework for additive runtime grammar updates.
GramDelta retains the Earley state for the committed prefix. When a new production arrives, it seeds that production at the historical waiters that can activate it, propagates only newly enabled deductions, and maps the updated state to exact next-token masks through compilation-free factorized projection: wide lexical continuations are resolved via grammar-independent vocabulary bitsets, while boundary-crossing multi-byte tokens are verified over shared historical charts without recompiling grammar-bound caches.

GramDelta addresses this problem through three components:

\begin{itemize}
    \item
    We characterize the historical entry points of additive grammar
    updates and prove that charts preceding the earliest entry remain unchanged,
    enabling direct update seeding without replaying the generated prefix.

    \item
    We derive semi-naive delta rules for Earley Predict, Scan, and Complete that
    propagate only computation involving new facts, and prove that their fixed
    point exactly matches full reconstruction under the updated grammar.

    \item
    We design a compilation free factorized token projector that couples
    grammar-independent lexical complement bitsets with origin consistent trie
    traversal over shared historical charts, preserving exact multi-symbol token
    validity without per-update cache recompilation.
\end{itemize}

We evaluate GramDelta across multiple model families, grammar structures, and
runtime-update workloads.
GramDelta preserves parser-state and token-validity equivalence to full
reconstruction in all evaluated cases, adding negligible overhead without
updates while reducing median first-seen grammar-update latency by
$10.7$--$11.5\times$ and the additional end-to-end TPOT overhead of sustained
runtime grammar evolution by $\mathbf{53.1}\text{--}\mathbf{81.9}\%$ relative to
the strongest reconstruction baseline.

\section{Background and Related Work}

\subsection{Grammar-Constrained Decoding}

Grammar-constrained decoding restricts the candidate tokens at each generation
step so that the generated output remains compatible with a formal language \citep{willard2023efficientguidedgenerationlarge,koo2024automata}.
Given a grammar $G$ over byte alphabet $\Sigma$, the byte sequence $x_{<t}\in\Sigma^t$ induced by the current generated
prefix, and tokenizer vocabulary $\mathcal V = \mathcal V_{\mathrm{byte}}\sqcup\mathcal V_{\mathrm{stop}}$ (excluding empty-payload special tokens), let $b(v)\in\Sigma^+$ denote the byte
sequence of token $v\in\mathcal V_{\mathrm{byte}}$.
The token mask $M_t\in\{0,1\}^{|\mathcal V|}$ is defined as
\[
    M_t[v]=1
    \iff
    \begin{cases}
        \exists y\in\Sigma^*,\quad x_{<t}b(v)y\in L(G), & v\in\mathcal V_{\mathrm{byte}}, \\
        x_{<t}\in L(G), & v\in\mathcal V_{\mathrm{stop}}.
    \end{cases}
\]

The central runtime task is therefore to determine valid next tokens from the
current prefix.
Existing systems accelerate this operation through different combinations of
automata or parser state, tokenizer-aware preprocessing, trie traversal,
caching, and grammar compilation.
For example, automata-based methods precompute tokenizer--language
compatibility, XGrammar compiles grammar structure for efficient runtime mask
generation \citep{dong2025xgrammar,li2026xgrammar}, Pre$^3$ \citep{chen-etal-2025-pre3} exploits deterministic pushdown transitions for LR(1)
grammars, and Earley-based systems such as Formatron \citep{pmlr-v267-sun25v} and \texttt{llguidance} \citep{llguidance} construct
token validity from incrementally advanced parser state.
These approaches substantially reduce constrained-decoding overhead when the
relevant grammar is fixed or already available.

We instead study what happens when new grammar structure becomes available
only after the current prefix has already been processed under an earlier
grammar.

\subsection{Runtime Grammar Evolution and State Maintenance}

Recent constrained-decoding systems support dynamic structure at several
levels.
\texttt{llguidance} uses an Earley recognizer together with a derivative-based lexer
and tokenizer-trie traversal to compute token masks for a supplied grammar
\citep{llguidance}.
XGrammar-2 targets dynamic agentic workloads: TagDispatch switches among
grammar components during generation, while cross-grammar caching,
adaptive Earley token-mask caching, and just-in-time compilation reuse work
across related structures
\citep{li2026xgrammar}.
Decode-Time Grammars provide a complementary form of dynamism by instantiating
grammar fragments from an evolving runtime environment $\Gamma$, allowing the
constraint applied to later regions to depend on declarations already
generated
\citep{zhang2026decode}.
Other code-generation methods similarly derive context-sensitive constraints
from type, scope, or program state
\citep{mundler2025type,li2025correctness}.
Collectively, these methods make generation-time constraints more efficient to
construct, select, or execute.

A different dimension of change arises when a grammar delta $\Delta G$ is
introduced only after the prefix $x_{<t}$ has already been processed under the
current grammar $G$.
Such an update can change not only the constraint to be executed subsequently,
but also the parsing state induced by the \emph{same prefix} under
$G \cup \Delta G$.
Existing reuse mechanisms do not directly maintain this historical state
across grammar versions: XGrammar-2 reuses grammar substructures and token-mask
caches, \texttt{llguidance} advances its recognizer over committed tokens under
a supplied grammar, Pre$^3$ precomputes pushdown transitions for a fixed LR(1)
grammar, and Decode-Time Grammars instantiate fragments for the corresponding
later generation regions.
Under such a late grammar update, obtaining the state equivalent to processing
the same prefix under the updated grammar therefore still requires
re-establishing its parsing computation, repeating work already performed.

Parsing research provides closely related forms of incremental reuse.
Incremental parsing reuses parsing results across input changes~\citep{Wagner_1998};
Reflective Grammars introduce scoped syntax extensions during
parsing~\citep{Stansifer_2011}; and DEXTER incrementally updates an LR goto graph as
productions are added or removed~\citep{Cazzola_2014}.
These approaches incrementally maintain either input-dependent parsing
results or grammar-dependent parser structure, but not prefix-dependent
historical state that must be revised by a later production addition while the
input itself remains fixed.
They therefore do not directly eliminate the historical recomputation induced
by the late-update setting above.

\section{GramDelta}
\label{sec:gramdelta}


Given a current grammar $G$, a committed prefix $x_{<t}$, and the parsing
state already materialized for that prefix, an additive grammar update
$\Delta G$ may become available at runtime.
Our objective is to obtain the parser state and next-token mask induced by the
updated grammar
\[
    G' = G \cup \Delta G
\]
without changing the committed prefix.

\begin{figure}[t]
    \centering
    \includegraphics[width=\linewidth]{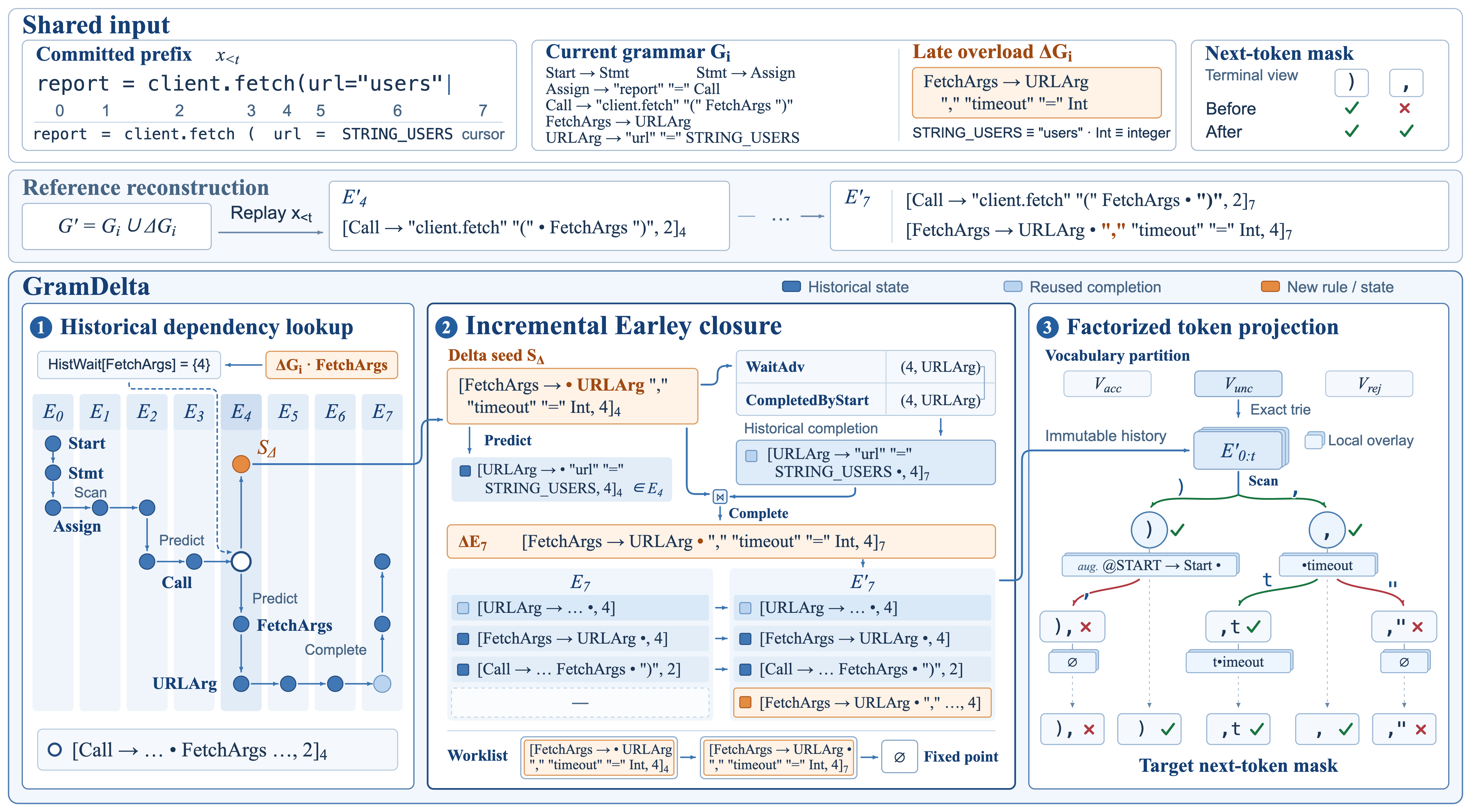}
    \caption{
Overview of \textsc{GramDelta}.
The top panel shows a committed code prefix, the current grammar, a late grammar update, and the resulting change in the valid next-token set.
The middle panel depicts the reference reconstruction path, which replays the committed prefix under the updated grammar.
The bottom panel shows the incremental \textsc{GramDelta} pipeline: historical dependency lookup finds the earliest affected waiting positions, incremental Earley closure propagates only newly enabled states while reusing historical completions, and factorized token projection computes the exact updated next-token mask.
The completed augmented item $\texttt{@START} \rightarrow \texttt{Start}\,\bullet$ indicates that the statement is fully accepted after consuming the closing parenthesis.
}
    \label{fig:overview}
\end{figure}

Figure~\ref{fig:overview} contrasts two ways of reaching this same target.
The reference reconstruction path builds the updated recognizer under $G'$,
replays the committed prefix from the beginning, and then projects the
reconstructed parser state to the next-token mask.
GramDelta instead reuses the materialized historical state and updates it in
three stages.

First, \emph{Historical Delta Localization} queries the persistent historical
waiting dependencies to identify the positions at which newly added
productions can enter the old Earley closure.
The earliest such position $j_\Delta$ gives a left boundary on historical
change, while all matching historical waiters contribute to the seed set
$S_\Delta$.
Second, \emph{Delta Earley Closure} propagates only newly enabled Predict,
Scan, and Complete deductions through an event-driven worklist, producing the
updated Earley state without replaying the prefix.
Third, \emph{Factorized Token Projection} combines grammar-independent lexical
complement bitsets with origin-consistent trie traversal over shared historical
charts to produce the exact next-token mask without recompiling grammar-bound
caches.

\subsection{Runtime Update Model}
\label{sec:update-model}

Let $G$ denote the current context-free grammar and $x_{<t}$ the byte sequence
induced by the committed generation prefix.
To ensure that active items admit finite terminal completions, deduction operates
over the productive subgrammar $G_{\mathrm{prod}}\subseteq G$ whose
right-hand-side nonterminals can derive terminal strings (tracked monotonically
via a productive-symbol registry; Appendix~\ref{app:earley-deduction}).
GramDelta retains the historical Earley state
\[
    E_{0:t}=(E_0,\ldots,E_t)
\]
where an item
\[
    [A\rightarrow\alpha\bullet\beta,o]_k\in E_k
\]
states that production $A\rightarrow\alpha\beta$ started at position $o$ and
that $\alpha$ has matched $x_{o:k}$.
We use the standard Predict, Scan, and Complete deductions \citep{Earley_1970}.

A previously unseen additive grammar update $\Delta G$ may become available
only after $x_{<t}$ has been committed, as occurs when newly declared
identifiers, inferred member signatures, or discovered schemas enter scope.
We consider production additions without deleting or modifying existing
productions
\[
    G'=G\cup\Delta G
    \qquad
    x_{<t}\text{ remains fixed}
\]
The post-update target is
\[
    E'_{0:t}
    =
    \operatorname{Earley}(G'_{\mathrm{prod}},x_{<t})
\]
together with the next-token mask $M'_t$ induced by this updated state.

Because productive membership and Earley deduction are monotone under
production addition
\[
    E_k\subseteq E'_k
    \qquad
    \Delta E_k=E'_k\setminus E_k
\]
GramDelta treats runtime grammar evolution as incremental maintenance of the
already materialized fixed point: it computes only the new facts
$\Delta E_{0:t}$ required to obtain $E'_{0:t}$, while preserving the committed
prefix and existing historical state.

\subsection{Historical Delta Localization}
\label{sec:localization}

When an update arrives, the old Earley closure has already been materialized.
GramDelta first determines where newly activated productions can enter this
historical computation.

Let $\Delta P_{\mathrm{act}}=P_{\mathrm{prod}}(G')\setminus P_{\mathrm{prod}}(G)$
denote all productions that become newly productive under $G'=G\cup\Delta G$,
comprising productive rules in $\Delta G$ together with any previously dormant
rules in $G$ whose right-hand sides become productive through $\Delta G$.
For a nonterminal $A$, define its historical waiting positions
\[
    \mathcal W_E(A)
    =
    \left\{
        k
        \;\middle|\;
        \exists [B\rightarrow\beta\bullet A\eta,o]_k\in E_k
    \right\}
\]
For each newly activated production $p:A\rightarrow\gamma\in\Delta P_{\mathrm{act}}$,
the corresponding delta seeds are
\[
    \operatorname{Seed}(p,E)
    =
    \left\{
        [A\rightarrow\bullet\gamma,k]_k
        \;\middle|\;
        k\in\mathcal W_E(A),\;
        [A\rightarrow\bullet\gamma,k]_k\notin E_k
    \right\}
\]
and
\[
    S_\Delta
    =
    \bigcup_{p\in\Delta P_{\mathrm{act}}}
    \operatorname{Seed}(p,E)
\]

$S_\Delta$ is exact rather than heuristic: it characterizes the entry points
through which newly activated productions can first leave the old Earley closure.

\textbf{Proposition 1 (First-new-fact characterization).}
Every
\[
    e\in E'_{0:t}\setminus E_{0:t}
\]
admits an Earley derivation whose first fact outside the old closure belongs
to $S_\Delta$.
Consequently
\[
    S_\Delta=\varnothing
    \quad\Longrightarrow\quad
    E'_{0:t}=E_{0:t}
\]

The proof and the corresponding historical left-boundary property are given
in Appendix~\ref{app:historical-localization}.

To avoid scanning historical charts when an update arrives, GramDelta
maintains a nonterminal-indexed relation $\texttt{HistWait}[A]$ during normal
decoding.
Whenever position $k$ contains an item waiting for $A$, $k$ is registered in
this relation.
A newly activated production $A\rightarrow\gamma$ can therefore instantiate
its seeds directly from $\texttt{HistWait}[A]$.

The retained dependency is semantic rather than representational:
historical chart rows may be organized into shared immutable pages, provided no
waiting or completion dependency that can be reactivated by a future grammar
addition is irreversibly discarded.

\subsection{Incremental Earley Closure}
\label{sec:incremental-closure}

The seed set $S_\Delta$ identifies where a grammar update first enters the
historical computation.
GramDelta then treats $E_{0:t}$ as a materialized Earley fixed point \citep{pereira1983} and
propagates only deductions newly enabled by these seeds \citep{Bancilhon_1986}.

\paragraph{Factorized Predict.}
Let $U$ encode active nonterminal waits
\[
    U(k,A)=1
    \iff
    E_k\text{ contains an item waiting for }A
\]
and let $V$ map nonterminals to production-entry states
\[
    V(A,q_p^0)=1
    \iff
    q_p^0\text{ is the entry state of }p:A\rightarrow\gamma\in P_{\mathrm{prod}}
\]
Predict can then be written as the factorized relation
\[
    \mathsf P=UV
\]
A grammar update extends $V$ by $\Delta V$, while incremental propagation may
introduce additional waits $\Delta U$.
The newly required Predict work is therefore
\[
    \Delta\mathsf P
    =
    U\Delta V
    \cup
    \Delta U V'
    \qquad
    V'=V\cup\Delta V
\]
The first term is exactly the historical seed relation from
Section~\ref{sec:localization}; the second propagates waits created during the
incremental closure.

\paragraph{Sparse Scan.}
Let $\mathsf S_a(q,q^+)$ denote a terminal transition from dotted state $q$
to its successor $q^+$ after consuming terminal or lexical symbol $a$.
Let $\Delta T_k$ be the sparse relation over dotted states and origins
defined by
\[
    \Delta T_k(q,o)=1
    \iff
    (q,o,k)\in\Delta E_k
    \text{ and }q\text{ expects a terminal or lexical transition}
\]
Because the committed input is unchanged, terminal-expecting facts already
present before the grammar update have already been scanned.
Only $\Delta T_k$ therefore requires new Scan work
\[
    \Delta E^{\mathrm{scan}}_{k+1}
    =
    \Delta T_k\,\mathsf S_{x_k}
\]

\paragraph{Origin-sensitive Complete.}
Complete joins a waiter for $B$ at position $j$ with a completed
$B$-constituent that starts at $j$.
GramDelta maintains
$\texttt{WaitAdv}[(j,B)]$ and
$\texttt{CompletedByStart}[(j,B)]$ for the two sides of this join.
A newly materialized waiter queries existing completions, while a newly
materialized completion queries existing waiters.
Whichever side arrives second therefore triggers the missing deduction
without rescanning historical charts.

\paragraph{Incremental fixed point.}
Every candidate fact passes chart-membership deduplication.
An unseen fact is inserted once, updates the corresponding sparse dependency
relations, and may trigger further Predict, Scan, or Complete consequences.
Propagation terminates when the worklist becomes empty.

The three delta operators above are combined in the event-driven procedure in
Algorithm~\ref{alg:gramdelta}.
A grammar update first inserts the exact historical seeds identified in
Section~\ref{sec:localization}.
Thereafter, every newly inserted fact triggers only the Predict, Scan, and
Complete consequences for which it supplies a previously missing premise.
Membership deduplication ensures that each new Earley fact is processed at most
once by the worklist.

\begin{algorithm}[t]
\caption{GramDelta recognizer update}
\label{alg:gramdelta}
\begin{algorithmic}[1]
\Require Persistent state $\mathcal R$, committed prefix $x_{<t}$,
         new productions $\Delta G$
\Ensure Updated Earley state

\State $\Delta P_{\mathrm{act}}\gets\Call{AppendProductions}{\Delta G}$
\ForAll{$p:A\rightarrow\gamma\in\Delta P_{\mathrm{act}}$}
    \ForAll{$k\in\texttt{HistWait}[A]$}
        \State \Call{Insert}{$[A\rightarrow\bullet\gamma,k]_k$}
    \EndFor
\EndFor

\While{\Call{WorklistNonEmpty}{}}
    \State $e\gets\Call{PopWorklist}{}$

    \If{$e$ waits for nonterminal $B$ at position $k$}
        \State register $k$ in $\texttt{HistWait}[B]$
               and $e$ in $\texttt{WaitAdv}[(k,B)]$
        \State \Call{Insert}{all productive Predict successors for $B$}
        \State \Call{Insert}{all joins with $\texttt{CompletedByStart}[(k,B)]$}

    \ElsIf{$e$ waits for terminal $\tau$ at $k<t$}
        \If{\Call{Matches}{$\tau,x_k$}}
            \State \Call{Insert}{the scanned successor of $e$}
        \EndIf

    \ElsIf{$e$ completes $B$ from $o$}
        \State register $e$ in $\texttt{CompletedByStart}[(o,B)]$
        \State join $e$ with $\texttt{WaitAdv}[(o,B)]$
    \EndIf
\EndWhile
\end{algorithmic}
\end{algorithm}

The two Complete directions in Algorithm~\ref{alg:gramdelta} are essential:
a new waiter must join previously materialized completions, while a new
completion must join previously materialized waiters.
Thus the procedure performs dependency-driven propagation from $S_\Delta$
rather than replaying the prefix from an affected position.

\textbf{Theorem 1 (Exact incremental closure).}
If
\[
    E_{0:t}
    =
    \operatorname{Earley}(G_{\mathrm{prod}},x_{<t})
\]
before the update and $G'=G\cup\Delta G$, then the state obtained at
termination satisfies
\[
    \widehat E_{0:t}
    =
    \operatorname{Earley}(G'_{\mathrm{prod}},x_{<t})
    =
    E'_{0:t}
\]

The complete delta derivation, proof, and event-driven procedure are given in
Appendices~\ref{app:incremental-closure} and~\ref{app:delta-update}.

\subsection{Factorized Token Projection}
\label{sec:token-projection}

Incremental Earley closure produces the updated historical parsing
state $E'_{0:t}$, from which GramDelta computes the updated next-token
mask $M'_t$. Since a token may span multiple grammar transitions and
trigger constituent completions originating at historical positions
$o\leq t$, an unchanged parser frontier does not imply that the
previous token mask remains valid. GramDelta therefore projects
tokens directly from the updated history without replaying the
committed prefix.

\paragraph{Vocabulary-level complement partitioning.}
GramDelta precomputes vocabulary-level symbol-containment sets
\[
    \mathcal I[\sigma]
    =
    \{v\in\mathcal V_{\mathrm{byte}}
      \mid \sigma\in b(v)\},
\]
together with the subset $\mathcal V_{\mathrm{closed}}$ of tokens
aligned to complete lexical atoms.
Based on the scannable terminal set $\Sigma_{\mathrm{first}}(E'_t)$
and the repeatable terminal set $\Sigma_{\mathrm{loop}}$ guaranteed
by active loops, the vocabulary is partitioned into directly
accepted, rejected, and uncertain tokens:
\[
\begin{aligned}
    \mathcal V_{\mathrm{acc}}
    &=
    \mathcal V_{\mathrm{closed}}
    \setminus
    \bigcup_{\sigma\notin\Sigma_{\mathrm{loop}}}
    \mathcal I[\sigma],\\
    \mathcal V_{\mathrm{rej}}
    &=
    \{v\in\mathcal V_{\mathrm{byte}}\setminus
    \mathcal V_{\mathrm{acc}}
    \mid b(v)_0\notin\Sigma_{\mathrm{first}}(E'_t)\},\\
    \mathcal V_{\mathrm{unc}}
    &=
    \mathcal V_{\mathrm{byte}}
    \setminus
    (\mathcal V_{\mathrm{acc}}\cup\mathcal V_{\mathrm{rej}}).
\end{aligned}
\]
When no active loop exists, $\mathcal V_{\mathrm{acc}}=\varnothing$.
This partition serves only as a fast filtering mechanism; all
uncertain candidates are handled by exact parser deduction.

\paragraph{Shared-history speculative trie deduction.}
For $\mathcal V_{\mathrm{unc}}$, GramDelta performs speculative
Earley deduction over the prefix trie $\mathcal T$ of candidate
token byte sequences. Each node $u$ at depth $d$ represents the
extension $x_{<t}\operatorname{str}(u)$ and maintains a speculative
parsing state at position $t+d$. The trie root directly reuses the
updated historical state $E'_{0:t}$. Advancing along an edge
$u\xrightarrow{\sigma}w$ applies Scan, Complete, and Predict under
$G'$, followed by the full Earley closure.

The key is to share the historical parsing base across candidate
branches. All branches treat $E'_{0:t}$ as immutable: speculative
completions can read historical waiters, but newly derived facts
are recorded only as branch-local additions at positions touched
along the active path. Consequently, common byte prefixes are
evaluated only once, while additions from different branches remain
isolated, avoiding redundant deduction and historical-state
contamination.

Let $E_{u_v}$ denote the speculative state at the trie leaf
corresponding to token $v$. A candidate is accepted if this state
contains an item expecting a further terminal or the completed
start-production item:
\[
    \Pi(E'_{0:t},\mathcal V_{\mathrm{unc}})
    =
    \{v\in\mathcal V_{\mathrm{unc}}
    \mid E_{u_v}\text{ contains a terminal-expecting item or }
    [S'\rightarrow S\bullet,0]\}.
\]

A stop token is valid only if the current state already contains
the completed start-production item:
\[
    M_t^{\mathrm{stop}}
    =
    \{v\in\mathcal V_{\mathrm{stop}}
    \mid [S'\rightarrow S\bullet,0]\in E'_t\}.
\]

The resulting token mask is
\[
    M'_t=
    \mathcal V_{\mathrm{acc}}
    \cup
    \Pi(E'_{0:t},\mathcal V_{\mathrm{unc}})
    \cup
    M_t^{\mathrm{stop}}.
\]

\textbf{Proposition 2 (Exact Factorized Token Projection).}
For any finite CFG $G'$ and committed prefix $x_{<t}$, if
$E'_{0:t}$ is the complete and correct historical Earley state under
$G'$, the fast-acceptance set satisfies its loop-preservation
invariant, and trie deduction performs complete Earley closure
after each byte, the resulting token mask is identical to that
obtained by full reconstruction under the updated grammar:
\[
    M'_t=M_t(G',x_{<t}).
\]

Proofs and persistent data-structure representations are provided
in Appendices~\ref{app:proof-projection} and
\ref{app:incremental-projection}.

\section{Experiments}
\label{sec:experiments}

We evaluate whether GramDelta reduces the cost of first-seen runtime grammar
evolution and whether this reduction translates to lower overhead in complete
serving requests.
All methods are evaluated within the same SGLang \cite{zheng2024sglang} serving stack under a common
late-update protocol.

\subsection{Experimental Setup}
\label{sec:experimental-setup}

\paragraph{Baselines.}
We compare GramDelta against internal and external baselines:
(i) \textbf{Fresh}, which reconstructs the recognizer inside the same native
Earley engine and replays the committed prefix, isolating incremental state
maintenance from implementation differences;
(ii) \textbf{XGrammar} \citep{dong2025xgrammar,li2026xgrammar},
evaluated both without cross-update compilation caching and with a persistent
warm compilation cache;
(iii) \textbf{\texttt{llguidance}} \citep{llguidance}, reusing
a persistent \texttt{ParserFactory} over idiomatic regex-lexer grammars.
Under our first-seen late-update protocol, reconstruction baselines instantiate
the updated matcher and reprocess the committed prefix, whereas GramDelta
applies the grammar delta directly to its persistent request-local recognizer
state.

\paragraph{Models and workloads.}
We evaluate Qwen3-8B \citep{yang2025qwen3}, Qwen3-32B, Granite-4.1-3B
\citep{granite2026}, and Granite-4.1-8B, covering multiple model scales and
tokenizer families.
Workloads span lightweight evolution, controlled local/recursive CFG updates
and compact progressive \texttt{BFCL} tool-calling traces \citep{patil2025berkeley}
($K=4\text{--}32$ updates), and sustained large-grammar evolution
(\texttt{CodeEvo}
declaration-driven code synthesis inspired by
\citet{mundler2025type,zhang2026decode}, with ${\sim}2.8\text{K}$ base rules and
$K=4\text{--}32$ updates), alongside static no-update controls ($K=0$).
Each grammar delta becomes visible only at its designated runtime position;
future grammar fragments are never preloaded.
All models execute on a single NVIDIA RTX PRO 6000 Blackwell GPU with a
single KV-cache prefill per request and no KV-cache reset upon grammar updates
(Appendix~\ref{app:experimental-details}).

\paragraph{Measurement protocol.}
For a first-seen grammar update, \emph{Update Latency} is the wall-clock
interval from the moment $\Delta G_i$ becomes visible to the constraint backend
until the token mask is available, excluding model forward execution
while including state maintenance or matcher construction, required prefix
reprocessing, and token projection.

\emph{TPOT} (Time Per Output Token) is measured over complete decoding requests
and therefore includes model execution, steady-state constrained decoding, and
runtime-update handling.
We report TPOT Overhead relative to unconstrained decoding:
\[
    \mathrm{TPOT\ Overhead}
    =
    \frac{
        \mathrm{TPOT}_{\mathrm{method}}
        -
        \mathrm{TPOT}_{\mathrm{unconstrained}}
    }{
        \mathrm{TPOT}_{\mathrm{unconstrained}}
    }
    \times 100\%.
\]
Headline comparisons use a shared-token protocol ($20\text{--}30$ paired runs
per cell) so that all backends follow the identical committed token sequence
and update chronology.

\subsection{Main Results}
\label{sec:main-results}

Table~\ref{tab:main-results} reports end-to-end GPU decoding results across all
four model configurations under the shared-token late-update protocol
(\texttt{BFCL}, $8\text{K}$ prefix, $K=32$ first-seen updates; $K=16$,
\texttt{CodeEvo}, and compact $512$-token traces exhibit consistent behavior in
Appendix~\ref{app:fullre}).
For every first-seen grammar update, all constrained backends process the
identical committed prefix and update sequence without resetting the model KV
cache.

\begin{table}[t]
\centering
\caption{
Main GPU decoding results across all four model configurations
(\texttt{BFCL}, $8\text{K}$ prefix, $K=32$ first-seen updates).
Update Latency is the median constraint-update time; TPOT Overhead is relative
to unconstrained decoding for the same model.
}
\label{tab:main-results}
\small
\renewcommand{\arraystretch}{1.15}
\setlength{\tabcolsep}{7pt}
\begin{tabular}{cc|ccc}
\hline
\textbf{Model} & \textbf{Method} &
\textbf{Update Lat. (ms)} $\downarrow$ &
\textbf{TPOT (ms/tok)} $\downarrow$ &
\textbf{TPOT Ovhd. (\%)} $\downarrow$ \\
\hline
\multirow{5}{*}{Qwen3-8B}
& Unconstrained & -- & 19.645 & 0.0 \\
& XGrammar & 500.680 & 124.953 & 536.1 \\
& llguidance & 49.437 & 27.078 & 37.8 \\
& Fresh & 42.066 & 28.855 & 46.9 \\
& \textbf{GramDelta} & \textbf{3.817} & \textbf{21.174} & \textbf{7.8} \\
\hline
\multirow{5}{*}{Qwen3-32B}
& Unconstrained & -- & 61.117 & 0.0 \\
& XGrammar & 526.198 & 163.645 & 167.8 \\
& llguidance & 51.570 & 65.603 & 7.3 \\
& Fresh & 41.921 & 70.624 & 15.6 \\
& \textbf{GramDelta} & \textbf{3.748} & \textbf{62.570} & \textbf{2.4} \\
\hline
\multirow{5}{*}{Granite-4.1-3B}
& Unconstrained & -- & 15.396 & 0.0 \\
& XGrammar & 550.876 & 133.650 & 768.1 \\
& llguidance & 49.942 & 23.884 & 55.1 \\
& Fresh & 41.326 & 25.458 & 65.4 \\
& \textbf{GramDelta} & \textbf{3.795} & \textbf{16.929} & \textbf{10.0} \\
\hline
\multirow{5}{*}{Granite-4.1-8B}
& Unconstrained & -- & 20.151 & 0.0 \\
& XGrammar & 562.078 & 139.269 & 591.1 \\
& llguidance & 53.271 & 28.138 & 39.6 \\
& Fresh & 42.947 & 29.955 & 48.7 \\
& \textbf{GramDelta} & \textbf{3.810} & \textbf{21.674} & \textbf{7.6} \\
\hline
\end{tabular}
\end{table}

\paragraph{First-seen update latency.}
Across all four models, GramDelta achieves
$3.748$--$3.817$\,ms median update latency at
$K=32$, reducing the cost by $10.89$--$11.27\times$
relative to same-engine Fresh reconstruction.
Unlike reconstruction baselines, it avoids replaying
the committed prefix or recompiling the updated grammar.

\paragraph{End-to-end decoding cost.}
GramDelta adds only $1.45$--$1.53$\,ms/token
over unconstrained decoding, corresponding to
$2.4$--$10.0\%$ TPOT overhead across the four models.
It reduces additional TPOT overhead by
$67.62$--$81.93\%$ relative to \texttt{llguidance},
the strongest reconstruction baseline on this metric.
Additional update counts and lightweight workloads
are reported in Appendix~\ref{app:fullre}.

\subsection{Ablation Study}
\label{sec:ablation}

Table~\ref{tab:ablation} isolates each algorithmic component of GramDelta within
the same native diagnostic build on Qwen3-8B across CPU microbenchmarks and
$8\text{K}, K=32$ GPU decoding.
Replacing same-engine \textbf{Fresh} reconstruction ($2.398$\,ms narrow CPU
update; $41.324$\,ms GPU update and $29.659$\,ms/tok TPOT) with semi-naive
\textbf{Delta Closure} eliminates committed-token replay, reducing GPU TPOT to
$21.233$\,ms/tok.
Adding \textbf{\texttt{HistWait} Localization} removes linear scans over
historical charts when seeding $S_\Delta$, further cutting $8\text{K}$ GPU
update latency from $6.316$\,ms to $3.757$\,ms ($1.68\times$ faster), while
static $K=0$ controls across all four models confirm that retaining
\texttt{HistWait} and completion indices changes TPOT by only $-0.13\%$ to
$+0.12\%$ (Appendix~\ref{app:fullre}).

\begin{table}[t]
\centering
\caption{
Component ablation on Qwen3-8B across narrow local CPU updates, wide open-string
projection, and $8\text{K}, K=32$ GPU decoding.
}
\label{tab:ablation}
\vspace{3pt}
\small
\renewcommand{\arraystretch}{1.15}
\setlength{\tabcolsep}{5pt}
\begin{tabular}{c|cc|cc}
\hline
\multirow{2}{*}{\textbf{Variant}} &
\textbf{Narrow Upd.} &
\textbf{Wide Proj.} &
\textbf{GPU Update} &
\textbf{GPU TPOT} \\
& \textbf{(ms)} $\downarrow$ &
\textbf{(ms)} $\downarrow$ &
\textbf{(ms)} $\downarrow$ &
\textbf{(ms/tok)} $\downarrow$ \\
\hline
Fresh (Reconstruct + Replay) & 2.398 & 0.038 & 41.324 & 29.659 \\
\hline
GramDelta w/o \texttt{HistWait} & 0.126 & 0.036 & 6.316 & 21.233 \\
GramDelta w/o Shared History & 13.771 & 0.752 & 193.368 & 242.196 \\
GramDelta w/o Complement Partition & 0.109 & 543.081 & 3.852 & 21.240 \\
\hline
\textbf{Full GramDelta} & \textbf{0.104} & \textbf{0.034} & \textbf{3.757} & \textbf{21.208} \\
\hline
\end{tabular}
\end{table}

Shared-History overlays are essential for efficient
speculative projection: disabling them increases
$8\text{K}$ GPU update latency from $3.757$ to
$193.368$\,ms. Complement Partitioning primarily
protects wide lexical states: removing it increases
wide-string projection latency from $0.034$ to
$543.081$\,ms, while barely affecting structural
update cost. Together, these components prevent
historical duplication and unnecessary vocabulary-wide
parser deduction.

\subsection{Scaling, Locality, and Discussion}
\label{sec:analysis}

\begin{figure}[t]
    \centering
    \includegraphics[width=\linewidth]{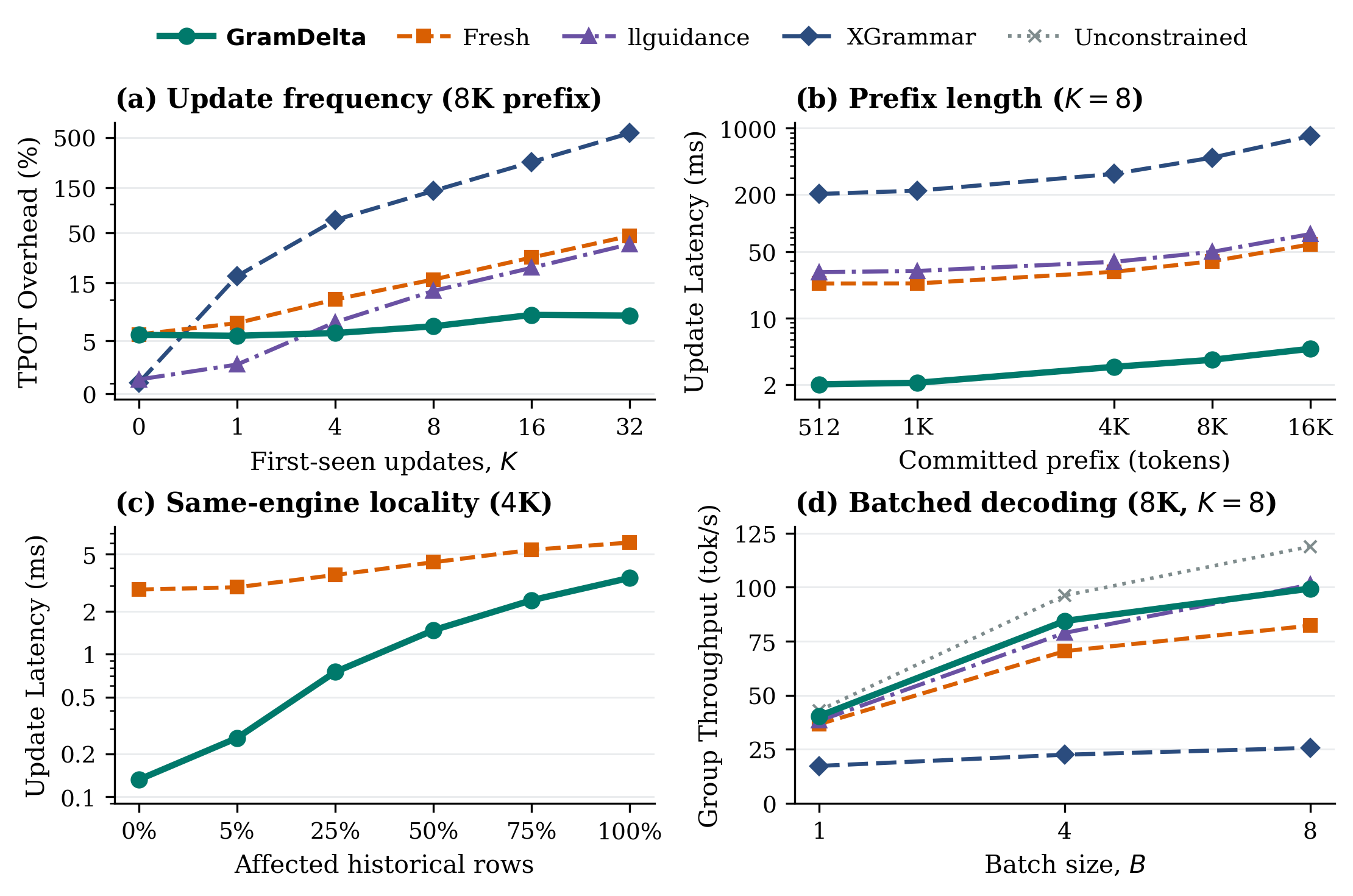}
\caption{
    Scaling and locality behavior on Qwen3-8B:
    (a) TPOT Overhead across first-seen update counts $K\in\{0,1,4,8,16,32\}$
    ($8\text{K}$ prefix);
    (b) Update Latency across committed prefix lengths ($512$--$16\text{K}$
    tokens, $K=8$);
    (c) same-engine Update Latency vs.\ affected historical chart rows
    ($4\text{K}$ prefix); and
    (d) fused batched decoding throughput across batch sizes $B\in\{1,4,8\}$
    ($8\text{K}, K=8$).
    }
    \label{fig:scaling}
\end{figure}

\paragraph{Scaling across update frequency, prefix
length, and batch size.}
Figure~\ref{fig:scaling} shows that GramDelta becomes
increasingly advantageous as runtime updates accumulate,
while precompiled XGrammar and \texttt{llguidance}
retain lower overhead in static or low-update regimes.
At $K=32$, GramDelta achieves $7.39\%$ TPOT overhead,
compared with $38.33\%$ for \texttt{llguidance},
$46.76\%$ for Fresh, and $557.47\%$ for XGrammar.
Its update latency increases only from $2.01$ to
$4.76$\,ms as the prefix grows from $512$ to
$16\text{K}$ tokens. Under fused batching, GramDelta
leads at smaller batch sizes and remains competitive
with \texttt{llguidance} at $B=8$.

\paragraph{Dependence on update locality.}
The benefit of incremental maintenance depends on
update locality. As affected historical rows increase
from $0\%$ to $100\%$, GramDelta's update latency
rises from $0.132$ to $3.414$\,ms, narrowing its
speedup over Fresh from $21.4\times$ to $1.77\times$.
In broad-history regimes, such as dense global LR(1)
completions or unbounded right recursion, automaton
reconstruction can outperform incremental Earley repair.
This affects performance rather than semantics:
GramDelta still produces the exact state and token mask.
Its persistent indices also incur modest memory overhead
(Appendix~\ref{app:fullre}).

\section{Conclusion}

We introduced GramDelta, an incremental framework for
constrained decoding under append-only runtime grammar
updates. By combining historical dependency localization,
semi-naive Earley closure, and exact factorized token
projection, GramDelta maintains the same parser state and
next-token mask as full reconstruction without replaying
the committed prefix. Its benefits depend on update locality:
broad changes may favor reconstruction, while supporting
deletions and replacements requires provenance-aware
retraction.

\bibliography{iclr2027_conference}

@inproceedings{geng2023grammar,
  title={Grammar-constrained decoding for structured NLP tasks without finetuning},
  author={Geng, Saibo and Josifoski, Martin and Peyrard, Maxime and West, Robert},
  booktitle={Proceedings of the 2023 Conference on Empirical Methods in Natural Language Processing},
  pages={10932--10952},
  year={2023}
}

@inproceedings{scholak-etal-2021-picard,
    title = "{PICARD}: Parsing Incrementally for Constrained Auto-Regressive Decoding from Language Models",
    author = "Scholak, Torsten  and
      Schucher, Nathan  and
      Bahdanau, Dzmitry",
    editor = "Moens, Marie-Francine  and
      Huang, Xuanjing  and
      Specia, Lucia  and
      Yih, Scott Wen-tau",
    booktitle = "Proceedings of the 2021 Conference on Empirical Methods in Natural Language Processing",
    month = nov,
    year = "2021",
    address = "Online and Punta Cana, Dominican Republic",
    publisher = "Association for Computational Linguistics",
    url = "https://aclanthology.org/2021.emnlp-main.779/",
    doi = "10.18653/v1/2021.emnlp-main.779",
    pages = "9895--9901"
}

@article{dong2025xgrammar,
  title={Xgrammar: Flexible and efficient structured generation engine for large language models},
  author={Dong, Yixin and Ruan, Charlie F and Cai, Yaxing and Xu, Ziyi and Zhao, Yilong and Lai, Ruihang and Chen, Tianqi},
  journal={Proceedings of Machine Learning and Systems},
  volume={7},
  year={2025}
}

@inproceedings{li2026xgrammar,
  title={XGrammar-2: Dynamic and Efficient Structured Generation Engine for Agentic LLMs},
  author={Li, Linzhang and Dong, Yixin and Wang, Guanjie and Xu, Ziyi and Jiang, Alexander and Chen, Tianqi},
  booktitle={Proceedings of the ACM Conference on AI and Agentic Systems},
  pages={1009--1022},
  year={2026}
}

@inproceedings{
hua2026gramtoken,
title={Gram2Token: Enabling Run-time {GPU}-Native Grammar-Constrained Decoding for {LLM}s},
author={Hantao Hua and Jiming Su and Hao Tang and Yiping Yao and Feng Zhu},
booktitle={Forty-third International Conference on Machine Learning},
year={2026},
url="https://openreview.net/forum?id=h3K23f6tLU"
}

@article{mundler2025type,
  title={Type-constrained code generation with language models},
  author={M{\"u}ndler, Niels and He, Jingxuan and Wang, Hao and Sen, Koushik and Song, Dawn and Vechev, Martin},
  journal={Proceedings of the ACM on Programming Languages},
  volume={9},
  number={PLDI},
  pages={601--626},
  year={2025},
  publisher={ACM New York, NY, USA}
}

@article{zhang2026decode,
  title={Decode-Time Grammars: Constrained LLM Generation over a Refinement Order of Grammar Fragments},
  author={Zhang, Shuoming and Xu, Ruiyuan and Li, Haofeng and Yu, Qiuchu and Zhang, Yangyu and Xia, Chunwei and Feng, Xiaobing and Wang, Chenxi and Cui, Huimin and Zhao, Jiacheng},
  journal={arXiv preprint arXiv:2607.18357},
  year={2026}
}

@article{li2025correctness,
  title={Correctness-guaranteed code generation via constrained decoding},
  author={Li, Lingxiao and Rahili, Salar and Zhao, Yiwei},
  journal={arXiv preprint arXiv:2508.15866},
  year={2025}
}

@article{koo2024automata,
  title={Automata-based constraints for language model decoding},
  author={Koo, Terry and Liu, Frederick and He, Luheng},
  journal={arXiv preprint arXiv:2407.08103},
  year={2024}
}

@misc{willard2023efficientguidedgenerationlarge,
      title={Efficient Guided Generation for Large Language Models}, 
      author={Brandon T. Willard and Rémi Louf},
      year={2023},
      eprint={2307.09702},
      archivePrefix={arXiv},
      primaryClass={cs.CL},
      url={https://arxiv.org/abs/2307.09702}, 
}

@inproceedings{chen-etal-2025-pre3,
    title = "Pre$^3$: Enabling Deterministic Pushdown Automata for Faster Structured {LLM} Generation",
    author = "Chen, Junyi  and
      Bai, Shihao  and
      Wang, Zaijun  and
      Wu, Siyu  and
      Du, Chuheng  and
      Yang, Hailong  and
      Gong, Ruihao  and
      Liu, Shengzhong  and
      Wu, Fan  and
      Chen, Guihai",
    editor = "Che, Wanxiang  and
      Nabende, Joyce  and
      Shutova, Ekaterina  and
      Pilehvar, Mohammad Taher",
    booktitle = "Proceedings of the 63rd Annual Meeting of the Association for Computational Linguistics (Volume 1: Long Papers)",
    month = jul,
    year = "2025",
    address = "Vienna, Austria",
    publisher = "Association for Computational Linguistics",
    url = "https://aclanthology.org/2025.acl-long.551/",
    doi = "10.18653/v1/2025.acl-long.551",
    pages = "11253--11267",
    ISBN = "979-8-89176-251-0"
}

@InProceedings{pmlr-v267-sun25v,
  title = 	 {Earley-Driven Dynamic Pruning for Efficient Structured Decoding},
  author =       {Sun, Xintong and Wei, Chi and Tian, Minghao and Ni, Shiwen},
  booktitle = 	 {Proceedings of the 42nd International Conference on Machine Learning},
  pages = 	 {57773--57789},
  year = 	 {2025},
  editor = 	 {Singh, Aarti and Fazel, Maryam and Hsu, Daniel and Lacoste-Julien, Simon and Berkenkamp, Felix and Maharaj, Tegan and Wagstaff, Kiri and Zhu, Jerry},
  volume = 	 {267},
  series = 	 {Proceedings of Machine Learning Research},
  month = 	 {13--19 Jul},
  publisher =    {PMLR},
  url = 	 {https://proceedings.mlr.press/v267/sun25v.html}
}

@article{Wagner_1998, title={Efficient and flexible incremental parsing}, volume={20}, ISSN={1558-4593}, url={http://dx.doi.org/10.1145/293677.293678}, DOI={10.1145/293677.293678}, number={5}, journal={ACM Transactions on Programming Languages and Systems}, publisher={Association for Computing Machinery (ACM)}, author={Wagner, Tim A. and Graham, Susan L.}, year={1998}, month=Sept, pages={980–1013} }

@inproceedings{Stansifer_2011, series={LDTA ’11}, title={Parsing reflective grammars}, url={http://dx.doi.org/10.1145/1988783.1988793}, DOI={10.1145/1988783.1988793}, booktitle={Proceedings of the Eleventh Workshop on Language Descriptions, Tools and Applications}, publisher={ACM}, author={Stansifer, Paul and Wand, Mitchell}, year={2011}, month=Mar, pages={1–7}, collection={LDTA ’11} }

@article{Cazzola_2014, title={On the incremental growth and shrinkage of LR goto-graphs}, volume={51}, ISSN={1432-0525}, url={http://dx.doi.org/10.1007/s00236-014-0201-2}, DOI={10.1007/s00236-014-0201-2}, number={7}, journal={Acta Informatica}, publisher={Springer Science and Business Media LLC}, author={Cazzola, Walter and Vacchi, Edoardo}, year={2014}, month=June, pages={419–447} }

@misc{llguidance,
  author       = {{Guidance AI}},
  title        = {{LLGuidance}: Super-fast Structured Outputs},
  howpublished = {\url{https://github.com/guidance-ai/llguidance}},
  year         = {2025},
  note         = {Software repository}
}

@article{Earley_1970, title={An efficient context-free parsing algorithm}, volume={13}, ISSN={1557-7317}, url={http://dx.doi.org/10.1145/362007.362035}, DOI={10.1145/362007.362035}, number={2}, journal={Communications of the ACM}, publisher={Association for Computing Machinery (ACM)}, author={Earley, Jay}, year={1970}, month=Feb, pages={94–102} }

@article{pereira1983,
  title={Parsing as deduction. ACL proceedings},
  author={Pereira, FCN and Warren, DHD},
  year={1983},
  journal={21st Annual meeting}
}

@inproceedings{Bancilhon_1986, series={SIGMOD ’86}, title={An amateur’s introduction to recursive query processing strategies}, url={http://dx.doi.org/10.1145/16894.16859}, DOI={10.1145/16894.16859}, booktitle={Proceedings of the 1986 ACM SIGMOD international conference on Management of data  - SIGMOD ’86}, publisher={ACM Press}, author={Bancilhon, Francois and Ramakrishnan, Raghu}, year={1986}, pages={16–52}, collection={SIGMOD ’86} }

@article{zheng2024sglang,
  title={Sglang: Efficient execution of structured language model programs},
  author={Zheng, Lianmin and Yin, Liangsheng and Xie, Zhiqiang and Sun, Chuyue and Huang, Jeff and Yu, Cody H and Cao, Shiyi and Kozyrakis, Christos and Stoica, Ion and Gonzalez, Joseph E and others},
  journal={Advances in neural information processing systems},
  volume={37},
  pages={62557--62583},
  year={2024}
}

@inproceedings{patil2025berkeley,
  title={The berkeley function calling leaderboard (bfcl): From tool use to agentic evaluation of large language models},
  author={Patil, Shishir G and Mao, Huanzhi and Yan, Fanjia and Ji, Charlie Cheng-Jie and Suresh, Vishnu and Stoica, Ion and Gonzalez, Joseph E},
  booktitle={Forty-second International Conference on Machine Learning}
}

@article{yang2025qwen3,
  title={Qwen3 technical report},
  author={Yang, An and Li, Anfeng and Yang, Baosong and Zhang, Beichen and Hui, Binyuan and Zheng, Bo and Yu, Bowen and Gao, Chang and Huang, Chengen and Lv, Chenxu and others},
  journal={arXiv preprint arXiv:2505.09388},
  year={2025}
}

@misc{granite2026,
  author       = {{IBM Research}},
  title        = {Granite 4.1 Language Models},
  year         = {2026},
  howpublished = {\url{https://huggingface.co/blog/ibm-granite/granit-4-1}},
  note         = {Accessed: 2026-04-28}
}
\bibliographystyle{iclr2027_conference}

\appendix
\section{Formal Details and Proofs}
\subsection{Earley Deduction System}
\label{app:earley-deduction}

We represent the runtime constraint grammar as a context-free grammar
\[
    G=(N,\Sigma,P,S)
\]
where $N$ is the set of nonterminals, $\Sigma$ the terminal alphabet, $P$ the
production set, and $S$ the start symbol.
Given a committed input prefix
\[
    x_{<t}=x_0x_1\cdots x_{t-1}
\]
which is fixed after generation, the Earley recognizer maintains a chart
$E_k$ at each input position $k\in\{0,\ldots,t\}$.

An Earley item is written as
\[
    [A\rightarrow\alpha\bullet\beta,o]_k
\]
meaning that production $A\rightarrow\alpha\beta$ started at input position
$o$, and $\alpha$ has matched the interval $x_{o:k}$.
We refer to $o$ as the \emph{origin} and $k$ as the current chart position.
Charts use set semantics, so each item is materialized at most once.

For a fixed grammar and input, the Earley state is the least closure under the
following three standard inference rules.

\paragraph{Predict.}
If an item in $E_k$ expects a nonterminal $B$, every production of $B$
introduces an initial item at the same position:
\[
\frac{
    [A\rightarrow\alpha\bullet B\beta,o]_k
    \qquad
    (B\rightarrow\gamma)\in P
}{
    [B\rightarrow\bullet\gamma,k]_k
}
\]

\paragraph{Scan.}
If an item at position $k$ expects terminal $a$ and the current input symbol
matches $a$, the dot advances into the next chart:
\[
\frac{
    [A\rightarrow\alpha\bullet a\beta,o]_k
    \qquad
    x_k=a
}{
    [A\rightarrow\alpha a\bullet\beta,o]_{k+1}
}
\]
For lexical terminals, $x_k=a$ can be replaced by the corresponding lexical
matching relation.
The arguments below rely only on the fact that Scan advances monotonically
from $k$ to $k+1$, rather than on a particular lexical representation.

\paragraph{Complete.}
If a constituent $B$ that started at position $j$ completes at position $k$,
every item in $E_j$ waiting for $B$ can advance:
\[
\frac{
    [A\rightarrow\alpha\bullet B\beta,o]_j
    \qquad
    [B\rightarrow\gamma\bullet,j]_k
}{
    [A\rightarrow\alpha B\bullet\beta,o]_k
}
\]
Thus, Complete is keyed by $(j,B)$: the origin $j$ of the completed
constituent must coincide with the chart position containing its waiter.
This origin-sensitive dependency is used later to construct the incremental
completion indices.

We denote the least fixed point generated from the augmented start item
\[
    [S'\rightarrow\bullet S,0]_0
\]
as
\[
    E_{0:t}
    =
    \operatorname{Earley}(G,x_{<t})
\]
The prefix is accepted as a complete string when
\[
    [S'\rightarrow S\bullet,0]_t\in E_t
\]
For constrained decoding, however, we additionally retain incomplete states
in $E_t$ that admit further terminal continuations, as these states determine
next-token validity.

To guarantee that every active item admits a finite terminal completion,
deduction operates over the productive subgrammar
$G_{\mathrm{prod}}=(N_{\mathrm{prod}},\Sigma,P_{\mathrm{prod}},S)$, where
$N_{\mathrm{prod}}(G)=\{A\in N\mid \exists w\in\Sigma^*,\; A\Rightarrow^*_G w\}$
and
$P_{\mathrm{prod}}(G)=\{A\rightarrow\gamma\in P\mid \gamma\in(N_{\mathrm{prod}}(G)\cup\Sigma)^*\}$.
We consider additive grammar updates of the form
\[
    G'=(N',\Sigma',P\cup\Delta P,S)
\]
while keeping $x_{<t}$ unchanged.
Since adding productions cannot remove an existing derivation, both
$P_{\mathrm{prod}}(G)\subseteq P_{\mathrm{prod}}(G')$ and Earley closure are
monotone:
\[
    E_k\subseteq E'_k,
    \qquad
    E_{0:t}=\operatorname{Earley}(G_{\mathrm{prod}},x_{<t}),
    \qquad
    E'_{0:t}=\operatorname{Earley}(G'_{\mathrm{prod}},x_{<t})
\]
We can therefore write
\[
    E'_k
    =
    E_k\cup\Delta E_k,
    \qquad
    \Delta E_k=E'_k\setminus E_k
\]
and characterize the update entirely in terms of the new Earley facts induced
by the newly activated productions
$\Delta P_{\mathrm{act}}=P_{\mathrm{prod}}(G')\setminus P_{\mathrm{prod}}(G)$.

\subsection{Proofs for Historical Delta Localization}
\label{app:historical-localization}

This section provides the proofs omitted from
Section~\ref{sec:localization}.
We reuse the notation defined there.

\paragraph{Proof of the First-New-Fact Characterization.}
Consider a finite Earley derivation of any
\[
    f\in E'\setminus E
\]
and choose a new fact $g\notin E$ of minimum derivation depth.
By minimality, every Earley premise of $g$ already belongs to $E$.

If $g$ is produced by \emph{Scan}, its only premise belongs to $E$.
The grammar update does not modify the committed prefix, and $E$ is already
closed under Scan for this fixed input.
Hence the same deduction must already have produced $g$ in the old closure,
contradicting $g\notin E$.

If $g$ is produced by \emph{Complete}, both its waiting and completed premises
belong to $E$.
Since the old Earley state is closed under Complete, their conclusion must
also belong to $E$, yielding the same contradiction.

Thus $g$ must be produced by \emph{Predict}.
Let
\[
    g=[A\rightarrow\bullet\gamma,k]_k
\]
be derived from a waiter
\[
    [B\rightarrow\beta\bullet A\eta,o]_k\in E_k
\]
If $A\rightarrow\gamma\in P_{\mathrm{prod}}(G)$, the same Predict deduction was
already available in the old productive grammar, and therefore $g\in E$.
Hence
\[
    A\rightarrow\gamma\in\Delta P_{\mathrm{act}}
\]
Because the waiter occurs in $E_k$, we have
$k\in\mathcal W_E(A)$, so $g\in S_\Delta$.
Therefore every derivation that produces a new Earley fact must first leave
the old closure through one of the seeds defined in
Section~\ref{sec:localization}.

\paragraph{Historical left boundary.}
Assume $S_\Delta\neq\varnothing$ and define the earliest seeded chart position
\[
    j_\Delta
    =
    \min
    \left\{
        k
        \;\middle|\;
        [A\rightarrow\bullet\gamma,k]_k\in S_\Delta
        \text{ for some }A\rightarrow\gamma\in\Delta P_{\mathrm{act}}
    \right\}
\]

\textbf{Corollary 1 (Historical left boundary).}
No chart preceding $j_\Delta$ changes under the additive update
\[
    \forall k<j_\Delta
    \qquad
    E'_k=E_k
\]

\emph{Proof.}
Every new derivation begins from a seed in $S_\Delta$.
Predict preserves the current chart position, Scan advances to the next
position, and Complete produces its consequence at the end position of the
completed constituent.
None of the three Earley deductions can propagate a newly derived fact to a
chart earlier than the position at which the new derivation entered the old
closure.
Since $j_\Delta$ is the earliest such entry position, no new fact can occur in
any $E_k$ with $k<j_\Delta$.

\paragraph{Proof of the Empty-Seed No-Op.}
Assume $S_\Delta=\varnothing$ but $E'\setminus E\neq\varnothing$.
The First-New-Fact Characterization would require the derivation of some new
fact to contain a first new fact in $S_\Delta$, a contradiction.
Thus $E'\setminus E=\varnothing$.
Together with monotonicity under additive updates, $E\subseteq E'$, this gives
\[
    E'=E
\]

\paragraph{Why the earliest seed alone is insufficient.}
The position $j_\Delta$ is only a left boundary on historical change.
For every
\[
    k\in\mathcal W_E(A)
\]
if
\[
    [A\rightarrow\bullet\gamma,k]_k
\]
was not already present in $E$, a newly activated production
$A\rightarrow\gamma\in\Delta P_{\mathrm{act}}$ makes this item itself a member
of $E'\setminus E$ through Predict.
Exact full-chart equivalence therefore requires retaining all seeds, rather
than only the earliest one.

\subsection{Proofs for Incremental Earley Closure}
\label{app:incremental-closure}

This section provides the derivations and correctness arguments omitted from
Section~\ref{sec:incremental-closure}, using the relational notation
introduced there.

\paragraph{Incremental Predict.}
The Predict relation is written as
\[
    \mathsf P=UV
\]
where $U(k,A)$ records that position $k$ contains a waiter for $A$, and
$V(A,q_p^0)$ maps a nonterminal to the entry state of one of its productive
productions.
A grammar update and subsequent closure propagation yield
\[
    V'=V\cup\Delta V,
    \qquad
    U'=U\cup\Delta U
\]
Hence
\[
\begin{aligned}
    \mathsf P'
    &=U'V' \\
    &=(U\cup\Delta U)(V\cup\Delta V) \\
    &=UV
      \cup U\Delta V
      \cup \Delta U V
      \cup \Delta U\Delta V
\end{aligned}
\]
Since $UV=\mathsf P$ and
\[
    \Delta U V'
    =
    \Delta U V
    \cup
    \Delta U\Delta V
\]
every potentially new Predict conclusion is generated by
\[
    U\Delta V
    \cup
    \Delta U V'
\]
Set-membership checks remove conclusions that have already been materialized.
The first term corresponds to existing historical waiters activated by newly
activated productions, while the second propagates Predict from waiters that
themselves arise during the incremental closure.

\paragraph{Incremental Scan.}
The lexical Scan relation is unchanged by the grammar update.
If a Scan conclusion becomes new after the update, its terminal-expecting
premise must also be new; otherwise the same deduction would already have
been available in the old closure for the unchanged committed prefix.
Therefore,
\[
    \Delta E^{\mathrm{scan}}_{k+1}
    =
    \Delta T_k\,\mathsf S_{x_k}
\]
where $\mathsf S_{x_k}$ denotes the Scan relation induced by the fixed input
symbol $x_k$.

\paragraph{Incremental Complete.}
Let $W(j,B)$ denote items at position $j$ waiting for $B$, and let $C(j,B)$
denote completed $B$-constituents whose origin is $j$.
Old Complete conclusions arise from
\[
    J=W\Join C
\]
With
\[
    W'=W\cup\Delta W,
    \qquad
    C'=C\cup\Delta C
\]
all potentially new joins are generated by
\[
    \Delta J
    =
    W\Join\Delta C
    \;\cup\;
    \Delta W\Join C'
\]
The first term handles a newly completed constituent against existing waiters;
the second handles a newly materialized waiter against existing or newly
materialized completions.
Any duplicate candidate is removed by membership checking.
The keyed structures described in Section~\ref{sec:incremental-closure}
implement these two cases by joining whichever side arrives later with the
already-materialized facts sharing the same $(j,B)$ key.

\paragraph{Exactness of incremental closure.}
Let $\widehat E$ denote the fixed point obtained by starting from $S_\Delta$
and repeatedly applying the incremental Predict, Scan, and Complete rules
above.
Then
\[
    \widehat E
    =
    E'
    =
    \operatorname{Earley}(G'_{\mathrm{prod}},x_{<t})
\]

\emph{Soundness.}
Every seed is a valid Predict conclusion under $G'_{\mathrm{prod}}$.
Every subsequently materialized fact is produced by a standard Earley rule
from premises already valid under $G'_{\mathrm{prod}}$.
Hence
\[
    \widehat E\subseteq E'
\]

\emph{Completeness.}
Consider any
\[
    f\in E'\setminus E
\]
By the First-New-Fact Characterization, some derivation of $f$ first leaves
the old closure through a fact in $S_\Delta$, which is inserted into the
initial worklist.

Proceed inductively along the remainder of this derivation.
If a new fact is produced by Predict, it is either an initial
$U\Delta V$ seed or is triggered by a newly materialized waiter through
$\Delta U V'$.
If it is produced by Scan, its premise must be new and is therefore propagated
by the incremental Scan rule.
If it is produced by Complete, at least one of its two premises must be new;
otherwise the old closure would already contain the conclusion.
A new completion is joined through $W\Join\Delta C$, while a new waiter is
joined through $\Delta W\Join C'$.
In every case, the new conclusion is eventually materialized.

Thus every fact in $E'\setminus E$ is produced, giving
\[
    E'\subseteq\widehat E
\]
Together with soundness,
\[
    \widehat E=E'
\]

For a fixed prefix and finite grammar, the Earley item universe is finite.
Since each item enters the worklist only on its first insertion, the
incremental closure also terminates.

\subsection{Proofs for Factorized Token Projection}
\label{app:proof-projection}

We prove Proposition~2 from Section~\ref{sec:token-projection}.
Fix $G'$ and $x_{<t}$, and assume that
$E'_{0:t}$ is the complete and correct Earley history under
$G'_{\mathrm{prod}}$.
For a byte token $v$, its mask membership is defined by
\[
    M_t(G',x_{<t})[v]=1
    \iff
    \exists y\in\Sigma^*:
    x_{<t}b(v)y\in L(G').
\]
A stop token is valid iff $x_{<t}\in L(G')$.

\paragraph{Lemma 1: Soundness of the fast paths.}
Consider an active productive loop witness $\ell$ in $E'_t$,
with a repeatable terminal set $\Sigma_\ell$.
By the loop-preservation invariant, consuming any complete
lexical atom from $\Sigma_\ell$, followed by Earley closure,
re-establishes the corresponding loop item at the resulting
position.
The directly accepted set can therefore be expressed as
\[
    \mathcal V_{\mathrm{acc}}
    =
    \bigcup_{\ell\in\mathcal L_t}
    \{v\in\mathcal V_{\mathrm{closed}}
      \mid b(v)\in\Sigma_\ell^*\},
\]
where $\mathcal L_t$ is the set of certified active loop
witnesses, and $\mathcal V_{\mathrm{acc}}=\varnothing$
when no such witness exists.

For any $v\in\mathcal V_{\mathrm{acc}}$, choose a witnessing
loop $\ell$ and decompose $b(v)$ into complete lexical atoms.
Starting from its loop item, each atom can be consumed by
Scan, followed by the corresponding Complete and Earley
closure steps, to re-establish the loop item at the next
position. By induction over the atoms of $b(v)$, the resulting
chart retains a viable loop state after the entire token.
Since the loop witness is productive and belongs to a
start-reachable derivation in $G'_{\mathrm{prod}}$, this state
admits a finite completion to a sentence. Hence,
\[
    M_t(G',x_{<t})[v]=1.
\]

Conversely, for any $v\in\mathcal V_{\mathrm{rej}}$,
the condition
$b(v)_0\notin\Sigma_{\mathrm{first}}(E'_t)$
implies that no Earley item can scan its first symbol.
The resulting chart is empty, and no continuation can complete
a derivation. Therefore,
\[
    M_t(G',x_{<t})[v]=0.
\]

\paragraph{Lemma 2: Equivalence of speculative trie deduction.}
For a trie node $u$ of depth $d$, let
$s_u=\operatorname{str}(u)$.
We show by induction on $d$ that its speculative state satisfies
\[
    E_u
    =
    \operatorname{Earley}
    (G'_{\mathrm{prod}},x_{<t}s_u)_{t+d}.
\]

For the root node, this follows directly from the reuse of
the complete historical state $E'_{0:t}$.
For an edge $u\xrightarrow{\sigma}w$, assume the equality
holds at $u$.
Appending $\sigma$ and applying Scan, Complete, Predict,
and the full Earley closure therefore produces exactly the
chart obtained by parsing
$x_{<t}s_u\sigma$ from the beginning.

The shared-history representation preserves this equivalence:
all branches read the immutable historical charts, while
branch-specific deductions are maintained in overlays along
their respective root-to-node paths.
All facts required by subsequent Complete operations are
retrieved from the shared historical state or the active
branch's overlay. Thus, sibling branches cannot introduce
facts into one another's states, and sharing a common trie
prefix does not alter the result of Earley deduction.

Moreover, pruning descendants after full closure when a
node contains no terminal-expecting items is exact, since
every proper descendant must consume at least one additional
terminal, which is impossible from such a state.

\paragraph{Lemma 3: Exact token-validity characterization.}
For a fully evaluated trie leaf $u_v$ corresponding to
$v\in\mathcal V_{\mathrm{unc}}$, let
\[
    \operatorname{Live}(E_{u_v})
\]
hold iff its chart contains either a terminal-expecting
Earley item or the completed start-production item
$[S'\rightarrow S\bullet,0]$.

Because the trie deduction performs full Earley closure,
a terminal-expecting item exposes a possible next terminal.
Since the remaining grammar context is productive in
$G'_{\mathrm{prod}}$, it admits a finite terminal completion.
Conversely, any nonempty continuation leading to a sentence
must expose a terminal-expecting item after closure.
The empty continuation is possible exactly when the
completed start-production item is present.
Consequently,
\[
\begin{aligned}
    \exists y\in\Sigma^*:
    x_{<t}b(v)y\in L(G')
    \iff
    \operatorname{Live}(E_{u_v}).
\end{aligned}
\]

It follows that
\[
    \Pi(E'_{0:t},\mathcal V_{\mathrm{unc}})
    =
    \{v\in\mathcal V_{\mathrm{unc}}
      \mid \operatorname{Live}(E_{u_v})\}.
\]

For a stop token $v\in\mathcal V_{\mathrm{stop}}$,
\[
\begin{aligned}
    v\in M_t^{\mathrm{stop}}
    &\iff [S'\rightarrow S\bullet,0]\in E'_t\\
    &\iff x_{<t}\in L(G').
\end{aligned}
\]

\paragraph{Proof of Proposition~2.}
By Lemma~1, the fast-acceptance and fast-rejection sets
classify their respective tokens correctly.
By Lemmas~2 and~3, speculative trie deduction classifies
every uncertain token exactly as full Earley parsing would.
Stop tokens are accepted precisely when the committed prefix
is already a complete sentence.

Since
\[
    \mathcal V_{\mathrm{byte}}
    =
    \mathcal V_{\mathrm{acc}}
    \sqcup
    \mathcal V_{\mathrm{rej}}
    \sqcup
    \mathcal V_{\mathrm{unc}},
\]
and the byte-token and stop-token vocabularies are disjoint,
their union yields
\[
    M'_t=M_t(G',x_{<t}).
\]
This completes the proof.

\subsection{Complexity Analysis}
\label{app:complexity}

Let $t=|x_{<t}|$ be the committed prefix length and
$|Q|=\sum_{p\in P'}(1+|\mathrm{RHS}(p)|)$ be the number of dotted grammar states
under $G'$.
Because chart deduplication inserts each newly derived fact in
$\Delta E=\bigcup_{k=j_\Delta}^t\Delta E_k$ into the worklist at most once,
Algorithm~\ref{alg:gramdelta} runs in output-sensitive time
\[
    O\Bigl(
        |\Delta P_{\mathrm{act}}|
        +
        \sum_{(q,o)_k\in\Delta E}
        \bigl(
            d_{\max}
            +
            |\texttt{WaitAdv}[(o,\cdot)]|
            +
            |\texttt{CompletedByStart}[(k,\cdot)]|
        \bigr)
    \Bigr),
\]
where $d_{\max}$ is the maximum number of production alternatives per
nonterminal.
When $S_\Delta=\varnothing$, historical update work reduces to
$O(|\Delta P_{\mathrm{act}}|)$; when an update triggers global derivations from
the start of the prefix, worst-case closure time is $O(|Q|t^3)$ for arbitrary
ambiguous CFGs and $O(|Q|t)$ for grammars with a bounded number of active
origins per dotted state.
Retaining persistent charts and reverse indices requires $O(|Q|t^2)$ space in
the worst case and $O(|Q|t)$ space under bounded active origins, while the
vocabulary sets $\mathcal I[\sigma]$ and $\mathcal V_{\mathrm{closed}}$ occupy
$O(|\Sigma|\cdot|\mathcal V|)$ static memory independent of $t$ and $|Q|$.

\section{Algorithms and Implementation}
\label{app:implementation}

\subsection{Event-Driven Delta Update}
\label{app:delta-update}

Algorithm~\ref{alg:gramdelta} in the main text gives the complete recognizer
update procedure.
Here we describe the persistent representation used to implement its
operations.

For a production
\[
    p:A\rightarrow X_1\cdots X_m
\]
GramDelta assigns stable dotted-state identifiers
\[
    q_p^0,\ldots,q_p^m
\]
An Earley fact is stored as
\[
    (q_p^d,o,k)
\]
where $o$ is the origin position and $k$ is the current chart position.
Runtime grammar additions append new production and dotted-state metadata in
fixed-size pages without copying unaffected pages, and identifiers associated
with previously materialized facts are never renumbered.

In addition to request-local chart membership, GramDelta maintains three
reverse indices
\[
    \texttt{HistWait}[A]
\]
\[
    \texttt{WaitAdv}[(k,A)]
\]
\[
    \texttt{CompletedByStart}[(k,A)]
\]
$\texttt{HistWait}[A]$ stores the deduplicated chart positions containing at
least one item that waits for $A$.
$\texttt{WaitAdv}[(k,A)]$ stores the corresponding waiting items themselves,
while $\texttt{CompletedByStart}[(k,A)]$ stores completed
$A$-constituents whose origin is $k$.
The latter two structures implement the two directions of the incremental
Complete join: a new waiter queries already materialized completions, and a
new completion queries already materialized waiters.

All derived facts pass through a single \textsc{Insert} operation.
For a candidate fact $f$, \textsc{Insert} checks request-local chart
membership.
If $f$ is already present, no further work is performed.
Otherwise, $f$ is materialized and placed on the worklist exactly once.
The worklist processing in Algorithm~\ref{alg:gramdelta} subsequently updates
the relevant reverse indices and triggers Predict, Scan, or Complete
successors.

For a newly activated production $A\rightarrow\gamma$, seed construction
queries only $\texttt{HistWait}[A]$.
No historical chart scan is required on this path.
Newly produced waiters and completions then propagate through the persistent
indices until the worklist becomes empty.
Thus the implementation follows the dependency-driven update characterized in
Section~\ref{sec:incremental-closure} rather than replaying the committed
prefix from an affected position.

\subsection{Implementation Details for Factorized Token Projection}
\label{app:incremental-projection}

GramDelta shares a static byte-level tokenizer trie and precomputed
64-bit packed vocabulary bitsets across requests using the same
tokenizer. At tokenizer initialization, each byte
$\sigma\in\{0,\ldots,255\}$ is mapped to a token-containment bitset
$\mathcal I[\sigma]$, while $\mathcal V_{\mathrm{closed}}$ identifies
tokens whose byte sequences consist of complete UTF-8 codepoints.
Tokens containing incomplete multi-byte character fragments bypass
bulk loop acceptance and are routed to exact trie deduction.

For active loops satisfying the loop-preservation invariant,
GramDelta computes $\mathcal V_{\mathrm{acc}}$ using 64-bit bitwise
operations over the precomputed containment sets, avoiding
per-token parser deduction.

To avoid duplicating historical charts across speculative trie
branches, committed rows $E'_{0:t}$ are organized into immutable
16-row pages. Each branch shares the page directory in $O(1)$ work.
When speculative Complete inserts a new completion at a historical
origin $o\leq t$, a row-level copy-on-write overlay is allocated
on demand; speculative charts at positions beyond $t$ are maintained
independently by each branch. For short prefixes $t\leq16$, an
inline single-page path reduces additional overhead.

The fast-filtering and trie-deduction stages jointly populate a
single packed token mask $M'_t$ per request, avoiding per-update
token-mask cache recompilation while preserving exactness across
multi-byte token boundaries.

\section{Experimental Details}
\label{app:experimental-details}

\subsection{Workloads and System Setup}
\label{app:workloads-setup}

We evaluate two classes of runtime grammar workloads across lightweight and
sustained scales.

First, we construct controlled dynamic grammars covering recursive updates,
subgrammar extension, locality sweeps ($0\%$--$100\%$ affected historical rows),
and compatible LR(1) fixtures for CPU comparison with Pre$^3$.
These families exercise recursive historical propagation, dormant-rule
reactivation, and broad-history crossover regimes.

Second, we evaluate growing schema and declaration workloads: compact BFCL
tool-calling traces ($512$-token context, $K=4\text{--}32$) and sustained
large-grammar evolution, \texttt{BFCL} derived from $128$ official BFCL
tool schemas and \texttt{CodeEvo} declaration-driven synthesis
(${\sim}2.8\text{K}$ base productions, $4\text{K}$--$16\text{K}$ prefixes,
$K=1\text{--}32$ first-seen updates).
This progressive-reveal protocol is introduced by our evaluation and is not
part of the original BFCL benchmark.

We evaluate Qwen3-8B, Qwen3-32B, Granite-4.1-3B, and Granite-4.1-8B.
All models execute in BF16 on a single NVIDIA RTX PRO 6000 Blackwell GPU within
a unified in-process KV-cache decoding harness using identical model weights,
tokenizers, and single-prefill decoding configurations.

\subsection{First-Seen Update Protocol}
\label{app:first-seen-protocol}

Each trace starts from an initial grammar $G_0$ and contains a sequence of
runtime grammar deltas
\[
    \Delta G_1,\ldots,\Delta G_m
\]
Before event $i$, a backend has access only to the current grammar
$G_{i-1}$.
Future grammar fragments are not registered or compiled in advance.
At the reveal event, $\Delta G_i$ becomes visible for the first time and the
active grammar becomes
\[
    G_i = G_{i-1}\cup\Delta G_i
\]
Updates occur after the current token has been committed, so all methods
process the same grammar change against the same historical prefix $x_{<t}$.

GramDelta applies $\Delta G_i$ directly to its persistent request-local
recognizer state.
For same-engine Fresh, XGrammar, \texttt{llguidance}, and Pre$^3$, we
instantiate the backend state corresponding to $G_i$ and reprocess the same
committed prefix.
This reconstruct-and-replay path describes their handling under our
first-seen late-update protocol rather than a general characterization of
those systems.

Update Latency is measured from the time $\Delta G_i$ becomes visible to the
constraint backend until the updated token mask is available for the next
decoding step.
Model forward execution is excluded from this event-level interval, while
constraint-state maintenance, required prefix reprocessing, and token
projection are included.
Each backend is independently warmed up before formal measurement, without
exposing grammar fragments belonging to future formal events.

The main comparison uses a shared-token protocol so that all backends follow the
same committed token sequence.

\paragraph{Token-policy alignment.}
For GramDelta, Fresh, and XGrammar mask comparisons, we exclude empty-payload
special tokens and admit configured stop tokens only in accepting states.
On shared-token trajectories, \texttt{llguidance} executes its public
\texttt{no\_forcing} path (verified at zero qualification mask differences),
while native fast-forward decoding is reported separately in
Appendix~\ref{app:fullre}.

\section{Extended Experimental Results}
\label{app:fullre}

This appendix provides the complete cross-model and cross-workload measurements
referenced in Section~\ref{sec:experiments}, followed by an analysis of
cross-domain behavior, compiler caching, model-scale interactions, and resource
tradeoffs.
Table~\ref{tab:cross-model} reports four-model GPU results on progressive
tool-schema discovery (\texttt{BFCL}, $8\text{K}$ prefix,
$K\in\{16,32\}$), and Table~\ref{tab:codeevo-4model} reports the corresponding
four-model GPU results on declaration-driven code synthesis (\texttt{CodeEvo},
$8\text{K}$ prefix, $K=32$).
Table~\ref{tab:static-control}  summarize
four-model static no-update controls ($K=0$).
Table~\ref{tab:update-stage-breakdown} decomposes the first-seen
update path across the two workloads.

\begin{table}[htbp]
\centering
\caption{
Complete four-model GPU results on \texttt{BFCL} ($8\text{K}$ prefix,
$K\in\{16,32\}$) with $95\%$ bootstrap confidence
intervals and p95 update latency.
}
\label{tab:cross-model}
\vspace{3pt}
\scriptsize
\renewcommand{\arraystretch}{1.12}
\setlength{\tabcolsep}{3pt}
\begin{tabular}{ccc|cccc}
\hline
\multirow{2}{*}{$\mathbf{K}$} &
\multirow{2}{*}{\textbf{Model}} &
\multirow{2}{*}{\textbf{Method}} &
\textbf{Update p50 [CI]} &
\textbf{Upd.\ p95} &
\textbf{TPOT p50 [CI]} &
\textbf{TPOT Ovhd.\ [CI]} \\
& & &
\textbf{(ms)} $\downarrow$ &
\textbf{(ms)} $\downarrow$ &
\textbf{(ms/tok)} $\downarrow$ &
\textbf{(\%)} $\downarrow$ \\
\hline
\multirow{24}{*}{16}
& \multirow{6}{*}{Qwen3-8B}
  & Unconstrained & -- & -- & 19.41 [19.40, 19.42] & 0.0 \\
& & XGrammar & 493.01 [488.85, 495.61] & 518.79 & 79.51 [79.21, 79.73] & 309.6 [308.0, 310.7] \\
& & XGrammar-Cache & 636.64 [635.03, 639.35] & 716.12 & 98.08 [97.52, 98.68] & 405.3 [402.4, 408.3] \\
& & llguidance & 49.76 [48.97, 50.63] & 57.31 & 23.93 [23.75, 24.08] & 23.3 [22.3, 24.0] \\
& & Fresh & 40.09 [39.50, 40.59] & 149.43 & 25.22 [25.20, 25.47] & 29.9 [29.8, 31.2] \\
& & \textbf{GramDelta} & \textbf{3.67 [3.60, 3.77]} & \textbf{4.83} & \textbf{20.69 [20.68, 20.72]} & \textbf{6.6 [6.5, 6.7]} \\
\cline{2-7}
& \multirow{6}{*}{Qwen3-32B}
  & Unconstrained & -- & -- & 61.05 [61.04, 61.07] & 0.0 \\
& & XGrammar & 495.24 [493.26, 498.00] & 522.79 & 116.16 [115.49, 116.39] & 90.3 [89.2, 90.6] \\
& & XGrammar-Cache & 642.86 [638.61, 646.19] & 719.85 & 135.33 [134.70, 135.79] & 121.7 [120.6, 122.4] \\
& & llguidance & 51.07 [50.38, 51.54] & 56.87 & 63.66 [63.56, 63.73] & 4.3 [4.1, 4.4] \\
& & Fresh & 41.41 [40.83, 41.86] & 168.34 & 66.89 [66.82, 66.99] & 9.6 [9.4, 9.7] \\
& & \textbf{GramDelta} & \textbf{3.76 [3.65, 3.87]} & \textbf{4.89} & \textbf{62.39 [62.36, 62.41]} & \textbf{2.2 [2.2, 2.2]} \\
\cline{2-7}
& \multirow{6}{*}{Granite-4.1-3B}
  & Unconstrained & -- & -- & 15.84 [15.79, 15.93] & 0.0 \\
& & XGrammar & 538.66 [534.11, 541.77] & 566.45 & 81.86 [81.48, 82.63] & 416.8 [413.2, 421.4] \\
& & XGrammar-Cache & 731.39 [725.68, 735.98] & 790.13 & 106.63 [105.17, 107.49] & 573.2 [563.6, 579.5] \\
& & llguidance & 50.72 [49.98, 51.29] & 55.02 & 20.76 [20.68, 20.96] & 31.0 [30.4, 32.3] \\
& & Fresh & 41.26 [40.71, 41.95] & 139.46 & 22.05 [21.68, 22.32] & 39.2 [36.8, 40.9] \\
& & \textbf{GramDelta} & \textbf{3.73 [3.59, 3.84]} & \textbf{4.97} & \textbf{17.20 [17.08, 17.37]} & \textbf{8.6 [7.7, 9.5]} \\
\cline{2-7}
& \multirow{6}{*}{Granite-4.1-8B}
  & Unconstrained & -- & -- & 19.87 [19.84, 20.22] & 0.0 \\
& & XGrammar & 528.97 [525.50, 532.68] & 560.58 & 84.65 [83.90, 85.17] & 326.0 [317.2, 328.7] \\
& & XGrammar-Cache & 717.66 [713.84, 720.14] & 790.43 & 108.54 [108.25, 109.21] & 446.2 [437.1, 449.4] \\
& & llguidance & 52.18 [51.27, 52.82] & 57.07 & 24.60 [24.40, 24.88] & 23.8 [21.5, 25.1] \\
& & Fresh & 42.12 [41.54, 42.80] & 156.58 & 26.22 [26.02, 26.72] & 32.0 [29.8, 34.2] \\
& & \textbf{GramDelta} & \textbf{3.89 [3.73, 3.97]} & \textbf{5.01} & \textbf{21.53 [21.31, 21.71]} & \textbf{8.3 [6.6, 9.2]} \\
\hline
\multirow{24}{*}{32}
& \multirow{6}{*}{Qwen3-8B}
  & Unconstrained & -- & -- & 19.65 [19.50, 19.73] & 0.0 \\
& & XGrammar & 500.68 [495.20, 506.55] & 576.22 & 124.95 [124.30, 127.10] & 536.1 [531.3, 548.3] \\
& & XGrammar-Cache & 636.70 [630.20, 644.25] & 749.03 & 156.51 [155.76, 157.65] & 696.7 [691.2, 705.4] \\
& & llguidance & 49.44 [48.57, 50.79] & 58.60 & 27.08 [26.80, 27.69] & 37.8 [36.4, 41.1] \\
& & Fresh & 42.07 [40.74, 44.14] & 142.63 & 28.86 [28.51, 29.40] & 46.9 [44.9, 50.0] \\
& & \textbf{GramDelta} & \textbf{3.82 [3.68, 3.90]} & \textbf{4.93} & \textbf{21.17 [21.13, 21.30]} & \textbf{7.8 [7.3, 8.7]} \\
\cline{2-7}
& \multirow{6}{*}{Qwen3-32B}
  & Unconstrained & -- & -- & 61.12 [61.11, 61.13] & 0.0 \\
& & XGrammar & 526.20 [524.03, 528.24] & 564.90 & 163.65 [163.39, 164.03] & 167.8 [167.3, 168.4] \\
& & XGrammar-Cache & 665.23 [662.82, 668.24] & 741.65 & 196.18 [195.71, 196.93] & 221.0 [220.2, 222.2] \\
& & llguidance & 51.57 [51.12, 52.05] & 57.08 & 65.60 [65.47, 65.68] & 7.3 [7.1, 7.5] \\
& & Fresh & 41.92 [41.41, 42.45] & 144.73 & 70.62 [70.25, 70.87] & 15.6 [14.9, 16.0] \\
& & \textbf{GramDelta} & \textbf{3.75 [3.66, 3.82]} & \textbf{4.87} & \textbf{62.57 [62.56, 62.60]} & \textbf{2.4 [2.3, 2.4]} \\
\cline{2-7}
& \multirow{6}{*}{Granite-4.1-3B}
  & Unconstrained & -- & -- & 15.40 [15.33, 15.51] & 0.0 \\
& & XGrammar & 550.88 [547.78, 554.50] & 594.93 & 133.65 [133.16, 134.20] & 768.1 [762.0, 773.1] \\
& & XGrammar-Cache & 738.40 [735.73, 740.64] & 816.81 & 176.52 [175.58, 176.90] & 1046.6 [1038.2, 1051.7] \\
& & llguidance & 49.94 [49.52, 50.39] & 54.37 & 23.88 [23.82, 24.09] & 55.1 [54.0, 56.8] \\
& & Fresh & 41.33 [41.03, 41.81] & 137.40 & 25.46 [25.22, 25.51] & 65.4 [63.5, 66.0] \\
& & \textbf{GramDelta} & \textbf{3.80 [3.71, 3.89]} & \textbf{4.94} & \textbf{16.93 [16.85, 17.01]} & \textbf{10.0 [9.3, 10.6]} \\
\cline{2-7}
& \multirow{6}{*}{Granite-4.1-8B}
  & Unconstrained & -- & -- & 20.15 [19.87, 20.28] & 0.0 \\
& & XGrammar & 562.08 [557.11, 565.88] & 610.07 & 139.27 [138.48, 140.58] & 591.1 [583.8, 606.1] \\
& & XGrammar-Cache & 750.82 [744.82, 758.24] & 835.19 & 183.20 [180.44, 184.45] & 809.1 [790.2, 827.5] \\
& & llguidance & 53.27 [52.37, 54.10] & 59.22 & 28.14 [27.92, 28.47] & 39.6 [38.1, 42.8] \\
& & Fresh & 42.95 [42.17, 43.83] & 138.66 & 29.96 [29.43, 30.06] & 48.7 [45.8, 51.0] \\
& & \textbf{GramDelta} & \textbf{3.81 [3.77, 3.91]} & \textbf{5.06} & \textbf{21.67 [21.39, 21.83]} & \textbf{7.6 [6.4, 8.9]} \\
\hline
\end{tabular}
\end{table}

\begin{table}[htbp]
\centering
\caption{
Complete four-model GPU results on declaration-driven code synthesis
(\texttt{CodeEvo}, $8\text{K}$ prefix, $K=32$) with
$95\%$ bootstrap confidence intervals and p95 update latency.
}
\label{tab:codeevo-4model}
\vspace{3pt}
\scriptsize
\renewcommand{\arraystretch}{1.12}
\setlength{\tabcolsep}{3.5pt}
\begin{tabular}{cc|cccc}
\hline
\multirow{2}{*}{\textbf{Model}} &
\multirow{2}{*}{\textbf{Method}} &
\textbf{Update p50 [CI]} &
\textbf{Upd.\ p95} &
\textbf{TPOT p50 [CI]} &
\textbf{TPOT Ovhd.\ [CI]} \\
& &
\textbf{(ms)} $\downarrow$ &
\textbf{(ms)} $\downarrow$ &
\textbf{(ms/tok)} $\downarrow$ &
\textbf{(\%)} $\downarrow$ \\
\hline
\multirow{6}{*}{Qwen3-8B}
  & Unconstrained & -- & -- & 19.70 [19.42, 19.91] & 0.0 \\
  & XGrammar & 362.07 [359.35, 365.79] & 391.88 & 78.90 [78.54, 80.57] & 300.5 [297.5, 307.4] \\
  & XGrammar-Cache & 376.24 [373.15, 382.09] & 425.65 & 82.41 [81.90, 83.18] & 318.3 [313.3, 324.9] \\
  & llguidance & 42.40 [41.59, 44.77] & 50.43 & 24.19 [23.97, 24.79] & 22.8 [21.4, 25.9] \\
  & Fresh & 33.67 [32.95, 34.93] & 134.89 & 26.55 [26.38, 26.78] & 34.8 [33.3, 36.8] \\
  & \textbf{GramDelta} & \textbf{3.15 [3.07, 3.23]} & \textbf{3.71} & \textbf{20.74 [20.66, 20.81]} & \textbf{5.3 [4.2, 6.8]} \\
\hline
\multirow{6}{*}{Qwen3-32B}
  & Unconstrained & -- & -- & 61.56 [61.54, 61.58] & 0.0 \\
  & XGrammar & 388.44 [387.31, 389.59] & 409.47 & 118.63 [118.48, 118.87] & 92.7 [92.4, 93.1] \\
  & XGrammar-Cache & 405.72 [404.55, 407.25] & 442.58 & 122.30 [122.11, 122.86] & 98.7 [98.3, 99.6] \\
  & llguidance & 46.71 [46.16, 47.64] & 52.30 & 64.11 [64.04, 64.20] & 4.1 [4.0, 4.3] \\
  & Fresh & 35.56 [35.07, 36.03] & 147.60 & 68.87 [68.83, 69.00] & 11.9 [11.8, 12.1] \\
  & \textbf{GramDelta} & \textbf{3.09 [3.00, 3.16]} & \textbf{3.66} & \textbf{62.76 [62.72, 62.78]} & \textbf{1.9 [1.9, 2.0]} \\
\hline
\multirow{6}{*}{Granite-4.1-3B}
  & Unconstrained & -- & -- & 15.56 [15.43, 15.79] & 0.0 \\
  & XGrammar & 414.28 [411.65, 417.64] & 443.08 & 85.15 [84.64, 86.09] & 447.4 [440.2, 455.4] \\
  & XGrammar-Cache & 459.38 [457.00, 462.17] & 495.97 & 93.46 [92.41, 94.12] & 500.8 [492.9, 505.5] \\
  & llguidance & 46.18 [45.60, 46.69] & 50.99 & 21.62 [21.46, 21.84] & 39.0 [36.3, 40.7] \\
  & Fresh & 35.26 [34.86, 35.74] & 132.42 & 23.33 [22.97, 23.44] & 49.9 [47.0, 51.5] \\
  & \textbf{GramDelta} & \textbf{3.19 [3.10, 3.29]} & \textbf{3.90} & \textbf{17.07 [17.01, 17.24]} & \textbf{9.7 [8.5, 11.1]} \\
\hline
\multirow{6}{*}{Granite-4.1-8B}
  & Unconstrained & -- & -- & 20.08 [20.07, 20.09] & 0.0 \\
  & XGrammar & 410.04 [407.64, 412.43] & 433.84 & 87.79 [87.40, 88.70] & 337.3 [335.3, 341.8] \\
  & XGrammar-Cache & 455.36 [451.51, 459.12] & 493.08 & 96.22 [95.64, 96.75] & 379.3 [376.5, 381.9] \\
  & llguidance & 46.68 [46.19, 47.31] & 52.44 & 25.22 [25.14, 25.41] & 25.6 [25.2, 26.6] \\
  & Fresh & 35.42 [35.14, 35.75] & 132.07 & 27.68 [27.61, 27.87] & 37.9 [37.5, 38.8] \\
  & \textbf{GramDelta} & \textbf{3.31 [3.19, 3.36]} & \textbf{3.92} & \textbf{21.66 [21.64, 21.68]} & \textbf{7.9 [7.8, 8.0]} \\
\hline
\end{tabular}
\end{table}

\begin{table}[htbp]
\centering
\caption{
Static no-update GPU controls ($K=0$, $30$ paired runs per cell) reporting
median TPOT ($1\text{K}$ / $4\text{K}$, ms/tok) and paired index-retention tax.
}
\label{tab:static-control}
\vspace{3pt}
\footnotesize
\renewcommand{\arraystretch}{1.15}
\setlength{\tabcolsep}{3.5pt}
\begin{tabular}{c|ccccc|c}
\hline
\multirow{2}{*}{\textbf{Model}} &
\textbf{Uncon.} &
\textbf{GD (Idx On)} &
\textbf{GD (Idx Off)} &
\textbf{XGrammar} &
\textbf{llguidance} &
\textbf{Index Tax} \\
& \textbf{($1\text{K}$ / $4\text{K}$)} &
\textbf{($1\text{K}$ / $4\text{K}$)} &
\textbf{($1\text{K}$ / $4\text{K}$)} &
\textbf{($1\text{K}$ / $4\text{K}$)} &
\textbf{($1\text{K}$ / $4\text{K}$)} &
\textbf{($1\text{K}$ / $4\text{K}$)} \\
\hline
Qwen3-8B & 17.79 / 18.85 & 18.05 / 19.19 & 18.03 / 19.19 & 17.94 / 19.01 & 18.03 / 19.09 & $+0.03\%$ / $+0.02\%$ \\
Qwen3-32B & 57.10 / 59.50 & 57.45 / 59.96 & 57.44 / 59.96 & 57.32 / 59.71 & 57.42 / 59.83 & $+0.01\%$ / $+0.00\%$ \\
Granite-4.1-3B & 15.18 / 15.14 & 15.36 / 15.45 & 15.34 / 15.41 & 15.30 / 15.23 & 15.39 / 15.40 & $-0.13\%$ / $+0.12\%$ \\
Granite-4.1-8B & 18.02 / 19.29 & 18.26 / 19.64 & 18.25 / 19.63 & 18.15 / 19.44 & 18.25 / 19.54 & $-0.03\%$ / $-0.11\%$ \\
\hline
\end{tabular}
\end{table}

\begin{table}[htbp]
\centering
\caption{
Mean update-stage decomposition (ms) for Qwen3-8B on
\texttt{BFCL} and \texttt{CodeEvo}.
}
\label{tab:update-stage-breakdown}
\vspace{3pt}
\scriptsize
\renewcommand{\arraystretch}{1.12}
\setlength{\tabcolsep}{3pt}
\begin{tabular}{llccccc}
\hline
\multirow{2}{*}{\textbf{Workload}} &
\multirow{2}{*}{\textbf{Stage}} &
\multicolumn{5}{c}{\textbf{Method (ms)}} \\
\cline{3-7}
& &
\textbf{GramDelta} &
\textbf{Fresh} &
\textbf{\texttt{llguidance}} &
\textbf{XGrammar} &
\textbf{XGrammar-Cache} \\
\hline

\multirow{8}{*}{\texttt{BFCL}}
& Dispatch + host preparation
& 1.487 & 38.658 & 12.256 & 12.521 & 12.381 \\

& Backend construction/update
& 1.480 & 10.918 & 23.001 & 514.676 & 663.804 \\

& Prefix replay
& 0.000 & 7.235 & 20.655 & 1.927 & 2.330 \\

& Historical lookup/closure
& 0.001 & 0.000 & 0.000 & 0.000 & 0.000 \\

& Token projection
& 0.098 & 0.054 & 0.062 & 0.008 & 0.008 \\

& Post-projection processing/return
& 0.728 & 0.504 & 0.025 & 0.004 & 0.004 \\

& Residual
& 0.022 & 0.023 & 0.012 & 0.064 & 0.066 \\

& \textbf{Total}
& \textbf{3.816} & \textbf{57.391}
& \textbf{56.011} & \textbf{529.200} & \textbf{678.592} \\

\hline
\multirow{8}{*}{\texttt{CodeEvo}}
& Dispatch + host preparation
& 1.074 & 34.511 & 11.471 & 12.330 & 12.254 \\

& Backend construction/update
& 1.149 & 5.692 & 21.338 & 362.190 & 387.924 \\

& Prefix replay
& 0.000 & 4.602 & 13.761 & 1.389 & 1.673 \\

& Historical lookup/closure
& 0.004 & 0.000 & 0.000 & 0.000 & 0.000 \\

& Token projection
& 0.165 & 0.060 & 0.049 & 0.007 & 0.007 \\

& Post-projection processing/return
& 0.663 & 0.341 & 0.021 & 0.004 & 0.003 \\

& Residual
& 0.028 & 0.022 & 0.012 & 0.062 & 0.056 \\

& \textbf{Total}
& \textbf{3.083} & \textbf{45.228}
& \textbf{46.652} & \textbf{375.981} & \textbf{401.918} \\

\hline
\end{tabular}
\end{table}

\paragraph{Cross-domain consistency (\texttt{BFCL} vs.\ \texttt{CodeEvo}).}
Comparing Tables~\ref{tab:cross-model} and~\ref{tab:codeevo-4model} demonstrates
consistent rankings across progressive tool-schema discovery and
declaration-driven code synthesis.
Absolute update and compilation latencies are moderately lower on
\texttt{CodeEvo} than on \texttt{BFCL} (e.g., $3.09\text{--}3.31$\,ms vs.\
$3.75\text{--}3.89$\,ms for GramDelta and $362.07\text{--}414.28$\,ms vs.\
$500.68\text{--}562.08$\,ms for XGrammar at $K=32$), because newly declared
identifiers and typed member signatures attach to localized expression
nonterminals rather than introducing nested multi-argument JSON object schemas.
Across all four models on \texttt{CodeEvo}, same-engine Fresh remains the
fastest reconstruction baseline on single-event Update Latency
($33.67\text{--}35.56$\,ms) while \texttt{llguidance} achieves the lowest
reconstruction TPOT ($4.1\text{--}39.0\%$ overhead); GramDelta outperforms both
simultaneously, achieving $10.70\text{--}11.50\times$ faster median updates than
Fresh and $53.08\text{--}76.82\%$ lower additional TPOT overhead than
\texttt{llguidance} ($97.66\text{--}98.24\%$ lower than XGrammar).
Furthermore, Fresh exhibits a wide p95 update tail ($132.07\text{--}147.60$\,ms
on \texttt{CodeEvo} and $137.40\text{--}168.34$\,ms on \texttt{BFCL})
because updates later in generation replay longer prefixes, whereas GramDelta
bounds p95 update latency to $3.66\text{--}3.92$\,ms on \texttt{CodeEvo} and
$4.83\text{--}5.06$\,ms on \texttt{BFCL}.

\paragraph{Effect of persistent compiler caching.}
Across every model and workload in Tables~\ref{tab:cross-model}
and~\ref{tab:codeevo-4model}, enabling XGrammar's persistent cross-grammar
compiler cache (\texttt{XGrammar-Cache}) yields higher latency than uncached
XGrammar (e.g., $376.24\text{--}459.38$\,ms vs.\ $362.07\text{--}414.28$\,ms on
\texttt{CodeEvo}, and $636.70\text{--}750.82$\,ms vs.\
$500.68\text{--}562.08$\,ms on \texttt{BFCL} at $K=32$).
Under the first-seen protocol, each composite grammar
$G_i = G_{i-1}\cup\Delta G_i$ contains previously unseen productions and misses
full-grammar cache entries while still incurring canonical hashing and
cache-insertion overhead, confirming that cross-grammar caching benefits
repeated structures rather than first-seen runtime additions.

\paragraph{Update-stage decomposition.}
Table~\ref{tab:update-stage-breakdown} decomposes the first-seen
update path across the two workloads.
GramDelta completes updates in $3.816$\,ms on \texttt{BFCL}
and $3.083$\,ms on \texttt{CodeEvo}, without replaying the
committed prefix.
Its remaining cost is primarily distributed between dispatch
and host preparation ($1.487$/$1.074$\,ms) and backend
update ($1.480$/$1.149$\,ms), while token projection and
post-projection processing together account for only
$0.826$/$0.828$\,ms.
In contrast, Fresh and \texttt{llguidance} incur substantial
prefix replay costs ($4.602$--$20.655$\,ms), whereas backend
construction/update accounts for $96.3\%$--$97.3\%$ of
XGrammar's total update latency.
Enabling persistent compiler caching does not reduce this
first-seen construction cost.
These results indicate that GramDelta's advantage primarily
comes from avoiding committed-prefix replay and full backend
reconstruction, rather than merely accelerating token
projection.

\end{document}